\documentclass{article}

\usepackage{arxiv}

\usepackage[utf8]{inputenc} 
\usepackage[T1]{fontenc}    
\usepackage{hyperref}       
\usepackage{url}            
\usepackage{booktabs}       
\usepackage{longtable}
\usepackage{amsfonts}       
\usepackage{nicefrac}       
\usepackage{microtype}      
\usepackage{graphicx}
\usepackage{float}
\usepackage[justification=raggedright,singlelinecheck=false]{caption}
\usepackage[justification=raggedright,singlelinecheck=false]{caption}
\graphicspath{ {./images/} }

\title{Tokenization Tradeoffs in Structured EHR Foundation Models}

\author{
Lin Lawrence Guo \\
Child Health Evaluative Sciences \\
The Hospital for Sick Children, Toronto, Canada \\
\And
Santiago Eduardo Arciniegas \\
Child Health Evaluative Sciences \\
The Hospital for Sick Children, Toronto, Canada \\
\And
Joseph Jihyung Lee \\
Child Health Evaluative Sciences \\
The Hospital for Sick Children, Toronto, Canada \\
\And
Adam Paul Yan \\
Child Health Evaluative Sciences \\
The Hospital for Sick Children, Toronto, Canada \\
Division of Haematology/Oncology \\The Hospital for Sick Children, Toronto, Canada \\
\And
George Tomlinson \\
Department of Medicine \\ 
University Health Network, Toronto, Canada \\
\And
Jason Fries \\
Department of Biomedical Data Science \\ 
Division of Computational Medicine, \\ Department of Medicine \\
Stanford University, Palo Alto, United States \\
\And
Lillian Sung\thanks{Corresponding author. Email: lillian.sung@sickkids.ca} \\
Child Health Evaluative Sciences \\
The Hospital for Sick Children, Toronto, Canada \\
Division of Haematology/Oncology \\ 
The Hospital for Sick Children, Toronto, Canada \\
}

\begin{document}
\maketitle

\begin{abstract}
Foundation models for structured electronic health records (EHRs) are pretrained on longitudinal sequences of timestamped clinical events to learn adaptable patient representations. Tokenization -- how these timelines are converted into discrete model inputs -- determines what information is preserved, how efficiently it is encoded, and which relationships must be learned versus pre-computed. Yet the impact of tokenization design choices on downstream performance and computational efficiency remains largely unexplored. Here, we pretrained a transformer on pediatric EHR data under a factorial design, varying tokenization along event encoding, time encoding, and workflow annotation. We evaluated area-under-the-receiver-operating-characteristic curve across 74 clinical prediction tasks. Joint event encoding and positional time encoding outperformed their alternatives (73/74 and 71/74 tasks) while requiring 39.5\% and 9.6\% fewer pretraining floating-point operations, respectively. Targeted ablations traced the joint encoding advantage to local binding efficiency, that is, code-attribute pairs are combined into single tokens, rather than split across tokens that the model must learn to associate during pretraining. External evaluation on an adult intensive care unit cohort demonstrated that this advantage generalizes despite substantial vocabulary mismatch, while temporal and workflow effects remain institution-specific. These results establish tokenization as a tractable lever for improving both the performance and efficiency of EHR foundation models.
\end{abstract}

\section{Introduction}
Foundation models are transforming artificial intelligence (AI) in healthcare by shifting development from bespoke, single-purpose models to reusable, adaptable backbones.\cite{bommasani2021,moor2023,wu2025,fu2025,singhal2025} Foundation models for structured electronic health records (EHR), trained on longitudinal sequences of timestamped clinical events such as diagnoses, procedures, medications, and laboratory results, have demonstrated improved prediction performance,\cite{steinberg2021,steinberg2023motor} robustness to distribution shifts,\cite{guo2023shift,guo2024multicenter,lemmon2023selfsupervised} and sample efficiency\cite{wornow2023neurips} in downstream clinical tasks. As these models advance toward broader clinical adoption,\cite{wornow2023shaky} understanding the design decisions that shape their learned representations becomes critical for guiding principled model development.

An important yet largely understudied step in training EHR foundation models is tokenization: converting a patient's clinical timeline into a sequence of discrete tokens. Because clinical events combine categorical codes, continuous measurements, and temporal relationships, the tokenization strategy determines what information is preserved in the token sequence (e.g., code-attribute associations, temporal intervals), how much of the patient's history fits within a fixed context window or compute budget, and which relationships the model must discover from data versus pre-computed during tokenization. Tokenization choices are fixed at pretraining time and propagate to every downstream application, making even modest effects consequential.

Three tokenization design axes recur across existing structured EHR foundation models. First, event encoding determines how individual clinical events are represented as tokens. An elevated serum glucose measurement, for example, may be encoded as a single composite token that fuses the code with its value (joint encoding),\cite{steinberg2021,kim2025pretrained} or decomposed into separate tokens that the model processes sequentially or aggregates prior to input (factorized encoding).\cite{renc2024zeroshot,waxler2025generative,hur2024genhpf,alattrach2025rethinking} Second, time encoding determines how temporal information is conveyed: whether through dedicated time interval tokens consuming sequence positions,\cite{renc2024zeroshot,pang2025cehrxgpt,pang2021cehrbert} positional encodings calibrated to temporal intervals,\cite{steinberg2021} or implicitly through sequence order alone.\cite{wornow2024contextclues,kraljevic2024foresight,rasmy2021medbert} Third, the scope of clinical context determines whether to represent clinical actions as single events (e.g., a finalized laboratory result) or as multi-step workflow sequences capturing process stages such as order placement, specimen collection, and result finalization. These choices jointly determine the vocabulary size, sequence length, and pretraining compute requirements of a model, and may further affect downstream task performance and cross-institutional generalizability. Yet these axes have been adopted without controlled comparison, hindering robust best practices.\cite{guo2026systematic}

Encoding clinical structure at tokenization time versus requiring the model to learn it from data reflects a fundamental trade-off in data-limited settings. EHR foundation models are pretrained on orders of magnitude fewer tokens than frontier language models,\cite{wornow2023shaky,guo2026systematic} limiting the signal available to learn complex associations. Binding at tokenization time hard-codes compositional structure but expands the vocabulary and increases parameters in the embedding layer. Factorized encoding preserves compositionality and a compact vocabulary but requires the model to learn code-value associations from pretraining data. For instance, joint event encoding pre-computes the association between a clinical code and its measured value within a single token, whereas factorized encoding requires the model to learn that the same value token carries different clinical meaning depending on which code it follows. This is a form of the binding problem in neural networks,\cite{greff2020binding} applied locally to adjacent tokens that must be interpreted together, which we refer to as local binding efficiency. Whether this trade-off yields measurable performance differences has not been well established for EHR foundation models.

Here, we quantify how tokenization design choices shape EHR foundation model performance and efficiency. Using a $2\times2\times2$ factorial experiment, we estimate the independent effects of event encoding, time encoding, and workflow annotations across 74 clinical prediction tasks spanning six clinical domains. We complemented this analysis with targeted ablation experiments to probe mechanisms and external evaluations to assess cross-institutional generalizability. Models with a fixed transformer architecture were pretrained on pediatric EHR data and evaluated locally and on an external adult intensive care cohort.\cite{johnson2023mimiciv} Our key findings are as follows:

\begin{enumerate}
\item Joint event encoding and positional time encoding improved downstream discrimination while requiring less pretraining compute, indicating that encoding more clinical structure at tokenization time improves both performance and efficiency.
\item Ablations traced the joint encoding advantage to local binding efficiency, whereby code-attribute pairs are pre-computed as single tokens during tokenization rather than learned across tokens from limited pretraining data.
\item Positional time encodings calibrated to temporal intervals showed modest improvement over sequence order alone, while dedicated time tokens degraded performance, indicating that how time is encoded matters more than whether it is encoded.
\item Benefits tied to institution-specific properties showed limited transfer: workflow annotations reflect local practice, and positional time encodings calibrated to one patient population did not generalize to populations with different age distributions.
\end{enumerate}

\section{Methods}

\subsection{Hospital Datasets}
This study used EHR data from The Hospital for Sick Children (SickKids), a tertiary pediatric hospital, as the primary development site, with data from Beth Israel Deaconess Medical Center (BIDMC), an adult academic medical center with intensive care unit admissions, used for external evaluation (MIMIC-IV).

The SickKids dataset was sourced from the SickKids Enterprise-wide Data in Azure Repository (SEDAR),\cite{guo2023sedar} which consolidates EHR data from SickKids' Epic Clarity database into a standardized schema of 20 clinically organized tables. EHR data were mapped to the Medical Event Data Standard (MEDS)\cite{arnrich2024meds,steinberg2024medsreader} format with clinical concepts standardized to Observational Medical Outcomes Partnership Common Data Model (OMOP CDM) ontologies. The MIMIC-IV dataset (version 1.0),\cite{johnson2023mimiciv} contains de-identified EHR data from patients admitted to the intensive care unit or emergency department at BIDMC between 2008 and 2019. MIMIC data were mapped to the OMOP CDM using code from the MIMIC project as part of Observational Health Data Sciences and Informatics\cite{ohdsi2021mimic} and subsequently converted to MEDS format. As part of MIMIC's de-identification process, patient timelines are shifted to an anchor year within a three-year window. To enable a consistent temporal splitting procedure across SickKids and MIMIC for pretraining and downstream evaluation, we deterministically assigned each patient a representative calendar year within their anchor group.

Use of SEDAR data for this study was approved by the Research Ethics Board (REB) at SickKids (REB number: 1000074527). Use of the MIMIC-IV dataset was approved under the oversight of the Institutional Review Boards (IRB) of BIDMC and the Massachusetts Institute of Technology (MIT). Access to MIMIC-IV data was granted through a credentialed data use agreement via PhysioNet.\cite{goldberger2000physiobank} The need for informed consent was waived by both organizations due to the retrospective nature of the project.

\subsection{Cohort Definition and Splitting}
The cohort selection process is summarized in Supplementary Figure S1. Pretraining cohorts were defined at the patient level. For SickKids, we included all patients in SEDAR at the time of the static data snapshot (May 7, 2025). Clinical events spanned from June 2, 2018 (EHR system go-live) through May 7, 2025. Patients were excluded if they died prior to the EHR system go-live date or had missing date of birth. For MIMIC, all patients in the dataset were included. Patients in each dataset were deterministically assigned to training (\textasciitilde90\%) and validation (\textasciitilde10\%) subsets. For SickKids, training patients contributed events occurring on or before May 31, 2023, and validation patients contributed events through May 31, 2024. For MIMIC, training events were included through December 31, 2016, with validation events included through December 31, 2017.

Downstream evaluation cohorts were defined at the admission level. For SickKids, we included inpatient admissions where age at admission was 28 days or older. For MIMIC, we included inpatient admissions where age at admission was 18 years or older. We excluded admissions in which death or discharge occurred on the day of admission. Admissions were assigned to training, validation, and test sets using temporal splits aligned with the pretraining cohorts.

Because pretraining cohorts were defined at the patient level and downstream evaluation cohorts at the admission level, patients in the downstream test set may have contributed historical clinical events to the pretraining corpus. This reflects realistic deployment, in which a foundation model trained on historical data is applied to new admissions from an overlapping patient population. No events occurring after the pretraining temporal cutoff were included in pretraining, and no events after the task-specific prediction time were used for representation extraction, preventing temporal information leakage while preserving realistic data utilization.

\subsection{Clinical Prediction Tasks}
We defined 74 clinical prediction tasks for SickKids, reflecting clinically relevant use cases developed under the Pediatric Real-world Evaluative Data sciences for Clinical Transformation (PREDICT)\cite{yan2025roadmap} program or based on prior SEDAR data requests. Tasks were grouped into six categories: blood bank transfusions (2 tasks), procedures (7 tasks), imaging (8 tasks), laboratory results (44 tasks, defined as abnormal high or low values relative to SickKids-specific reference ranges), medication administrations (10 tasks, categories defined using the American Hospital Formulary Service classification), and clinical outcomes (3 tasks: in-hospital mortality, long length of stay of 7 or more days, and 30-day readmission).

For MIMIC, we evaluated 13 tasks comprising an adapted subset of the SickKids tasks, including the same three clinical outcomes and 10 laboratory results tasks, with abnormal values defined using MIMIC-specific reference ranges.

The prediction time was set at midnight on the day of admission for all tasks except 30-day readmission, for which prediction time was set at midnight on the day before discharge. The prediction window extended until discharge for all tasks except for long length of stay and readmission, which used fixed windows of 7 days post-admission and 30 days post-discharge, respectively. For each task, admissions in which the outcome occurred between admission and the prediction time were excluded.

\subsection{EHR Tokenization Strategies}
\begin{figure}[H]
\centering
\includegraphics[width=\textwidth]{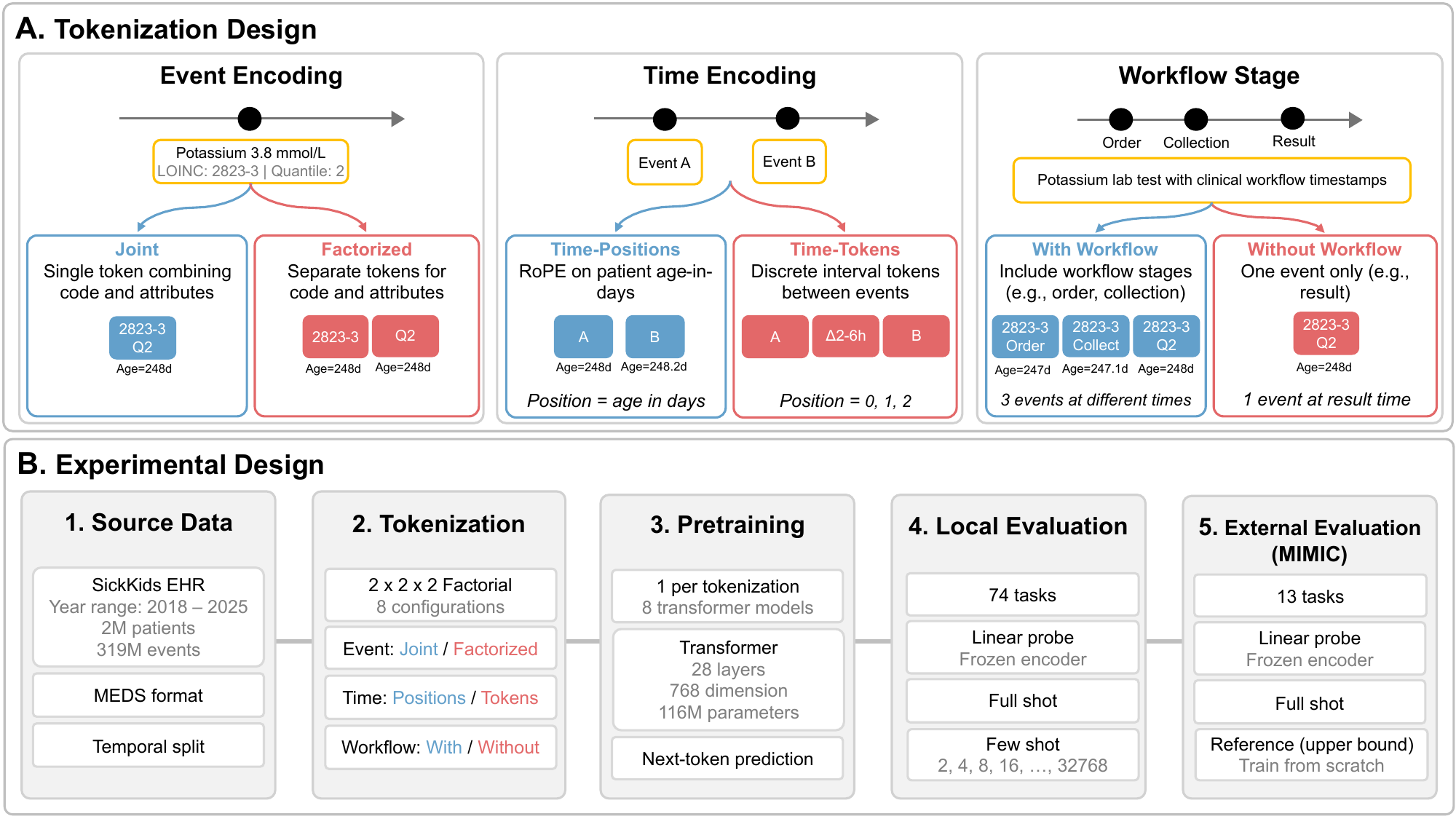}
\caption{Tokenization and experiment design. (A) Tokenization design choices. Event encoding determines how clinical events are represented: joint encoding creates a single token combining the event code and its attributes, while factorized encoding uses separate tokens for the code and each attribute. Time encoding determines how temporal information is captured: Time-Positions uses rotary positional embeddings (RoPE) on patient age-in-days, while Time-Tokens inserts discrete interval tokens between events with sequential integer positions. Workflow stage determines whether clinical workflow is included: with workflow, a single lab test generates separate events at order, collection, and result times; without workflow, only the result event is retained. (B) Experimental design. (1) Source data from SickKids including all patient events and workflow stages, (2) tokenization configurations based on the factorial design, (3) next-token-prediction pretraining of one transformer model per tokenization configuration, (4) local evaluation settings using 74 clinical prediction tasks, and (5) external evaluation using MIMIC across 13 tasks, with models trained from scratch on MIMIC serving as an upper-bound reference. Abbreviations: LOINC -- Logical Observation Identifiers Names and Codes; RoPE -- rotary positional embeddings; ORD -- order; COL -- collection; EHR -- electronic health records; MEDS -- medical event data standard.}
\label{fig:fig1}
\end{figure}

We evaluated three tokenization design axes in a $2 \times 2 \times 2$ factorial experiment (illustrated in Figure~\ref{fig:fig1}).

\subsubsection{Event Encoding}
We compared joint and factorized strategies for event encoding. In joint encoding, each clinical event was mapped to a single token representing the clinical concept and its associated attribute (e.g., a serum glucose measurement in the third quantile bin). In factorized encoding, events were decomposed into separate tokens: a base concept token followed by attribute tokens representing numeric quantile bins, categorical text values, and workflow stages, drawn from a small, shared vocabulary. For both strategies, numeric values were discretized into 10 decile bins per clinical concept using the pretraining dataset. Joint encoding produced one token per event, whereas factorized encoding produced two or more tokens per event, resulting in longer sequences and a smaller vocabulary.

\subsubsection{Time Encoding}
We compared two strategies for time encoding: Time-Positions and Time-Tokens. In Time-Positions, each token's positional index corresponded to the patient's age in days at the time of the event, with temporal relationships encoded via Rotary Positional Embeddings (RoPE).\cite{su2024roformer}

In Time-Tokens, temporal information was represented explicitly using discrete time-interval tokens inserted between consecutive clinical events, following prior work.\cite{renc2024zeroshot,waxler2025generative} Tokens were assigned sequential integer positions, and RoPE was applied to token order rather than patient age. We used 13 time-interval bins spanning 5 minutes to over 6 months (Supplementary Table~S1), and each patient sequence began with demographic tokens encoding binned age and biological sex. Insertion rules and bin boundaries are detailed in Supplementary Table~S1. Time-Tokens sequences were longer than Time-Positions due to the additional interval and demographic tokens.

\subsubsection{Workflow Stage Annotations}
We compared models with and without workflow stage annotations. In the without-workflow condition, each clinical action was represented as a single event at its primary timestamp (e.g., finalized result time for laboratory tests; start time for procedures and surgeries). In the with-workflow condition, clinical actions with multi-step workflows were represented using multiple events occurring at distinct timestamps corresponding to workflow stages (e.g., order placement, specimen collection, and result finalization for laboratory tests). This increased both information content and the sequence length of patient timelines. Event types without a multi-step workflow, such as diagnoses and flowsheet measurements, were represented identically in both conditions. Workflow-stage representations by clinical domain are provided in Supplementary Table~S2. Workflow stage annotations were available only in the SickKids dataset; equivalent information was not present in MIMIC.

The full factorial design yielded eight tokenization conditions for SickKids pretraining. A separate vocabulary was constructed for each condition from the pretraining dataset.

\subsection{Foundation Model Pretraining}
We adopted a decoder-only transformer\cite{vaswani2017attention,brown2020language} with 28 layers, a hidden dimension of 768, 12 attention heads, and an intermediate dimension of 1,152.\cite{warner2024modernbert} This configuration yields a fixed transformer backbone of 115.6 million parameters, which was held constant across all experiments (Supplementary Table~S3). The total parameter count varied across tokenization conditions (132M--170M parameters) due to differences in vocabulary size, which affect the size of both the input token embedding layer and the next-token prediction layer used during pretraining. We employed local attention with a window size of 128 tokens, with global attention every third layer.\cite{child2019sparse}

Eight models (one per factorial tokenization condition) were pretrained on the SickKids cohort using autoregressive next-token prediction. Each training batch was constructed using token-budget-based sequence packing, filling a fixed budget of 32,768 tokens corresponding to the model's maximum context window. Because patient timelines vary substantially in length, a batch could contain one long patient sequence or multiple shorter sequences concatenated together, with a minimum of one patient per batch. Causal attention masking enforced patient boundaries, preventing information leakage across patients within the same packed sequence. Each model was trained for five epochs without early stopping\cite{muennighoff2023scaling} on a single NVIDIA H100 or L40S GPU. To estimate compute cost, total pretraining floating-point operations (FLOPs) were calculated using the standard approximation of $6ND$,\cite{kaplan2020scaling} where $N$ is the number of non-embedding model parameters and $D$ is the total number of tokens processed during training. The transformer model was implemented using PyTorch version 2.7.\cite{paszke2019pytorch} Optimizer settings and additional training details are provided in Supplementary Table~S4.

To establish a domain-matched upper-bound reference, we additionally trained models from scratch on MIMIC using MIMIC-specific vocabularies, with identical architecture and training procedures. Because workflow stage annotations were not available in MIMIC, reference models were trained under a reduced $2 \times 2$ factorial design (event encoding $\times$ time encoding), all in the without-workflow condition.

\subsection{Ablation Experiments}
\subsubsection{Event Encoding Ablation}
Joint and factorized encoding differ along two axes that may independently affect downstream performance: token efficiency (joint encoding preserves longer clinical histories under fixed context windows) and local binding requirements (factorized encoding requires the model to learn that adjacent attribute tokens modify their preceding code token, whereas joint encoding encodes these associations at tokenization time). To disentangle these mechanisms, we conducted a $2 \times 4 \times 2$ factorial ablation. The first factor was encoding strategy (joint vs.~factorized). The second was information content at four levels: Code Only (clinical concept codes without attributes), +Attributes (codes with numeric and categorical value tokens, requiring local binding between adjacent tokens in the factorized condition), +Workflow (codes with workflow stage events at distinct timestamps, requiring no local binding in either condition), and Full (all information). The third was sequence length regime: Fixed-Length (32,768 tokens, the standard training budget) versus Fixed-Event (14,000 events, equalizing the number of clinical events observed regardless of encoding strategy). If token efficiency drives any performance difference between encoding strategies, the Fixed-Event regime (which removes the sequence length disparity) should reduce performance differences. If local binding requirements drive the difference, the +Attributes condition (which introduces binding demands) should reveal it, while the +Workflow condition (which adds information without binding demands) should not. This experiment used the same model architecture and training procedure and was conducted using Time-Positions as time encoding on the SickKids dataset. All models were pretrained under the Fixed-Length regime; the Fixed-Event condition was applied only during downstream representation extraction by truncating patient sequences to a fixed number of events.

\subsubsection{Time Encoding Ablation}
Motivated by recent findings that specialized time encodings often provide limited benefit over sequence order alone,\cite{alattrach2025rethinking,wornow2024contextclues} we sought to quantify the marginal contribution of explicit temporal information beyond event ordering. Using joint event encoding, we compared four time encoding strategies: (1) Order-Only, in which tokens received sequential integer positions with no temporal information; (2) Time-Positions; (3) Time-Tokens; and (4) Positions+Scalar, which supplements RoPE by overwriting two hidden-state dimensions with normalized age and age$^2$ at every transformer layer. The Positions+Scalar condition was included for completeness, as variants incorporating explicit continuous age features have been used in prior EHR foundation models.\cite{steinberg2021} Each strategy was evaluated with and without workflow stage annotations, yielding eight conditions. All models used the same architecture and training procedure on the SickKids dataset.

\subsection{Downstream Task Evaluation}
For each patient admission in the downstream evaluation cohorts, representations were extracted using the pretrained model for the corresponding tokenization condition. Specifically, for SickKids downstream tasks, representations were extracted using the SickKids-pretrained models. For MIMIC downstream tasks, representations were extracted using both SickKids-pretrained models (for transfer evaluation) and MIMIC-pretrained models (as domain-matched references).

Each model was applied autoregressively to the sequence of all clinical events occurring at or before the task-specific prediction time, up to the model's maximum context window of 32,768 tokens. The final hidden state (768-dimensional vector) corresponding to the last event in this sequence was used as the patient representation at prediction time. Pretrained model weights were frozen during representation extraction.

Downstream task performance was evaluated using the area under the receiver operating characteristic curve (AUROC). Performance was reported on the held-out temporal test set.

\subsubsection{Full-Shot Evaluation}
For full-shot evaluation, L2-regularized logistic regression models from Sci-kit learn\cite{pedregosa2011scikitlearn} were trained on patient representations extracted from all available task-specific training admissions. Regularization strength ($C \in \{1, 0.1, 0.01, 0.001, 0.0001\}$) was selected based on validation AUROC. Input features were standardized using a standard scaler fitted on the training set and applied to validation and test sets.

\subsubsection{Sample Efficiency Evaluation}
To assess sample efficiency, models were evaluated across shot sizes ($k$) ranging from 2 to 32,768 labeled training examples, in powers of two. For each shot size, 10 independent iterations were conducted. In each iteration, $k$ training examples were drawn from the task-specific training set, while the validation and test sets remained unchanged. Training subsets were constructed using balanced sampling, consisting of an equal number of positive and negative examples. For tasks with fewer than $k/2$ positive instances in the training set, all available positive instances were included in the training subset, with the remaining examples sampled from the negative class. The same L2-regularized logistic regression setup used for full-shot evaluation was applied in all sample efficiency experiments. Performance was reported as the mean across the 10 iterations for each shot size.

\subsection{External Evaluation on MIMIC}
Cross-institutional generalization was evaluated by applying the SickKids-pretrained models to MIMIC data using frozen model weights and transferred vocabularies. Only the downstream linear classifiers were trained on MIMIC data, following the same full-shot evaluation procedure. Transfer performance was compared to MIMIC reference models trained from scratch under the reduced factorial design. Out-of-vocabulary rates were quantified by applying each SickKids-derived vocabulary to MIMIC data and stratifying events by type (code-only, numeric, and categorical text attributes).

\subsection{Statistical Analysis}
We used linear mixed models (LMMs) to estimate the effects of tokenization choices on downstream task performance. In all models, the outcome variable was task-level AUROC, with each observation corresponding to a single task evaluated under a specific tokenization condition. Prediction task was included as a random intercept to account for non-independence of repeated evaluations within the same task. Model parameters were estimated using restricted maximum likelihood (REML). Confidence intervals and $p$-values were computed using Wald $t$-distribution approximation. For the full-shot and transfer factorial experiments, we fitted the following model:

\[
\mathrm{AUROC} \sim \mathrm{TimeEncoding} + \mathrm{EventEncoding} + \mathrm{Workflow} + (1|\mathrm{Task})
\]

The model estimates the independent contribution of each design axis. We excluded the interaction terms as the axes represent orthogonal design choices and the primary goal was to quantify the marginal effect of each. The event encoding ablation used a full factorial model (Encoding $\times$ Information $\times$ Regime) with task as a random intercept. For the time encoding ablation, a linear mixed model included time encoding strategy as a fixed effect with Order-Only as the reference level and task as a random intercept. For the sample efficiency analysis, the model was extended to capture interactions between tokenization choices and training set size:

\[
\mathrm{AUROC} \sim (\mathrm{TimeEncoding} + \mathrm{EventEncoding} + \mathrm{Workflow}) \times \log_2(\mathrm{ShotSize}_{\mathrm{centered}}) + (1|\mathrm{Task})
\]

Shot size was $\log_2$-transformed and centered at 32 examples, such that main effects correspond to performance at the 32-shot setting, and interaction terms capture how tokenization effects vary with training sample size.

\section{Results}
\subsection{Cohort Characteristics}
Pretraining and task cohort characteristics are summarized in Table~\ref{tab:table1}. The SickKids pretraining cohort consisted of 2,027,582 patients contributing 169 million clinical events with a median timeline duration of 1 day (IQR 0--224). The MIMIC pretraining cohort comprised 339,989 patients with 179 million events and a median timeline duration of 17 days (IQR 2--553). Task cohorts for downstream evaluation included 87,565 admissions across 51,242 pediatric patients (median age 7 years) at SickKids, and 58,513 admissions across 44,055 adult patients (median age 54 years) in MIMIC. Task-specific statistics including total admissions, patients, and prevalence for SickKids and MIMIC are summarized in Supplementary Table~S5.

\begin{table}[H]
\centering
\caption{Characteristics of pretraining and task cohorts}
\label{tab:table1}
\begin{tabular}{p{0.50\textwidth}cc}
\toprule
\textbf{Characteristic} & \textbf{SickKids} & \textbf{MIMIC} \\
\midrule
\textbf{Pretraining Cohort*} & & \\
Patients, n & 2,027,582 & 339,989 \\
Clinical events, n & 169,323,884 & 178,990,210 \\
Timeline duration in days, median (IQR) & 1 (0--224) & 17 (2--553) \\
Female sex, \% & 49.8 & 53.5 \\
\addlinespace
\textbf{Task Cohort**} & & \\
Patients, n & 51,242 & 44,055 \\
Admissions, n & 87,565 & 58,513 \\
Age at admission, median (IQR) & 7 (2--13) & 54 (34--70) \\
Length of stay in days, median (IQR) & 2 (1--5) & 4 (2--7) \\
Female sex, \% & 46.0 & 61.9 \\
Clinical prediction tasks, n*** & 74 & 13 \\
\bottomrule
\end{tabular}

\vspace{0.5em}
{\footnotesize *Pretraining cohorts include patients and events (after temporal splitting and not including workflow stages) used for self-supervised pretraining of the SickKids and the reference MIMIC foundation models. \\ ***Task cohorts include admissions used for downstream clinical prediction evaluation. The MIMIC cohort was also used for the external evaluation of the SickKids foundation model. \\ ****The clinical prediction tasks were identified from data requests to the SickKids team. Among the 74 identified tasks (Supplementary Table S5), 13 could be operationalized using data available in MIMIC. \\ Abbreviations: IQR -- interquartile range; SickKids -- The Hospital for Sick Children; MIMIC -- Medical Information Mart for Intensive Care.}
\end{table}

\subsection{Effect of Tokenization on Sequence Length and Pretraining Cost}
Across the eight tokenization configurations, vocabulary sizes ranged from 10,889 to 35,432 tokens and total pretraining tokens ranged from 169 to 479 million (Supplementary Table~S6). Factorized encoding reduced vocabulary size by 2.3--3.0 times relative to joint encoding but increased total tokens by 1.8--2.0 times under matched time and workflow settings. Time-Tokens increased total tokens by 6.9--16.4\% relative to Time-Positions. Total FLOPs ranged from $5.5 \times 10^{17}$ to $1.4 \times 10^{18}$, scaling with tokens seen during pretraining (Supplementary Table~S7).

\subsection{Tokenization Choices Independently Affect Task Performance}
\begin{figure}[H]
\centering
\includegraphics[width=\textwidth]{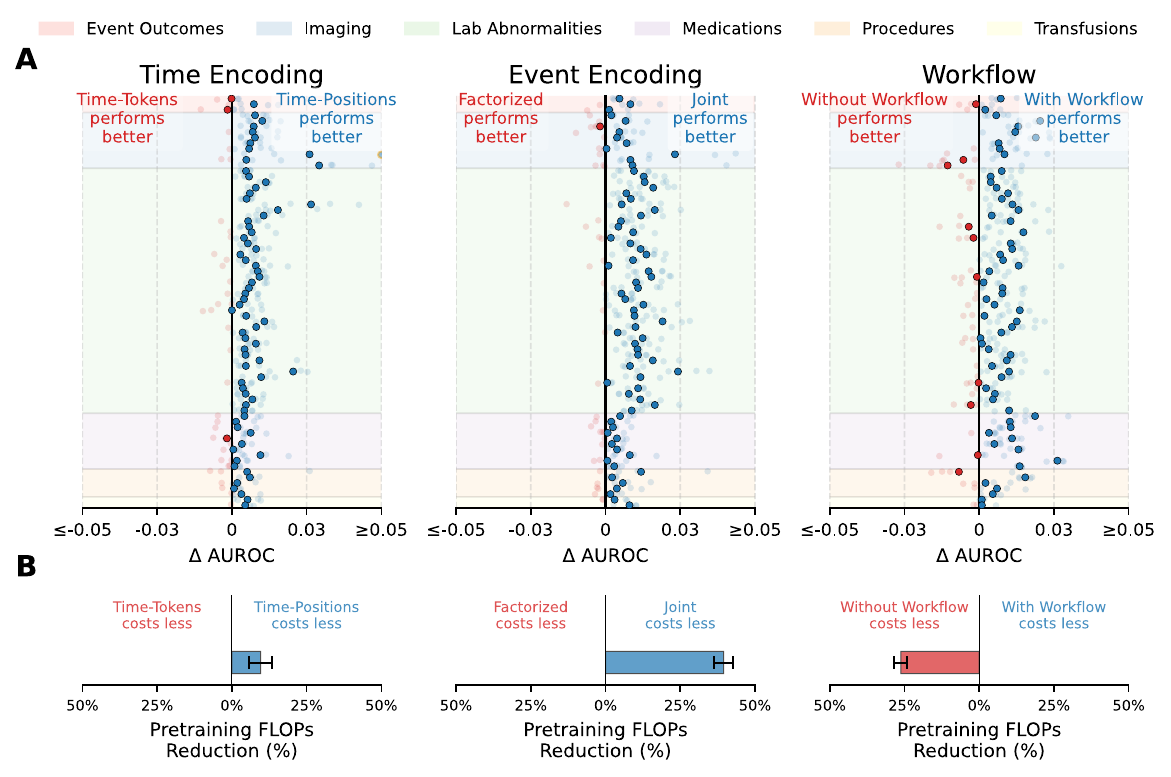}
\caption{Effect of tokenization design choices on task performance and pretraining cost. (A) Task-specific differences in AUROC between paired tokenization strategies across 74 clinical prediction tasks evaluated on the SickKids dataset. Each transparent point represents the AUROC difference for a single task under a specific experimental configuration. Opaque points denote the mean AUROC difference for each task, averaged across all other experimental factors. Background bands indicate task family. Absolute AUROCs by tokenization condition and task are reported in Supplementary Tables S8 and S9, respectively. (B) Relative difference in pretraining compute, measured as FLOPs between paired tokenization strategies. Bars indicate the mean percentage reduction across configurations, with error bars showing the range observed across experimental settings. Abbreviations: AUROC -- area under the receiver operating characteristic curve; FLOPs -- floating-point operations.}
\label{fig:fig2}
\end{figure}

All three tokenization design choices independently affected downstream task performance across 74 clinical prediction tasks (Figure~\ref{fig:fig2}A; Supplementary Tables~S8--S10). Joint encoding outperformed factorized encoding ($\beta = 0.008$ AUROC, 95\% CI [0.007, 0.009], $P < 0.001$), Time-Positions outperformed Time-Tokens ($\beta = 0.007$, 95\% CI [0.005, 0.008], $P < 0.001$), and including workflow stage annotations improved performance ($\beta = 0.007$, 95\% CI [0.006, 0.008], $P < 0.001$). When averaging across the other two design axes, these effects were consistent in direction across most tasks (joint: 73/74 [99\%]; Time-Positions: 71/74 [97\%]; workflow: 64/74 [86\%]). Notably, the better-performing strategies for event and time encoding also required less pretraining compute (joint: 39.5\% fewer FLOPs; Time-Positions: 9.6\% fewer FLOPs), while workflow inclusion presented a trade-off, with improved performance at 35.6\% greater compute cost (Figure~2B). These effects were stable across sample efficiency evaluation settings from 2 to 32,768 labeled examples (Supplementary Figure~S2; Supplementary Table~S11), with one exception: the time encoding advantage diminished at larger training set sizes (interaction $\beta = 0.001$, $P = 0.012$), while event encoding and workflow effects remained stable ($P$ not significant).

\subsection{Joint Encoding Advantage Reflects Local Binding Efficiency}
\begin{figure}[H]
\centering
\includegraphics[width=0.7\linewidth]{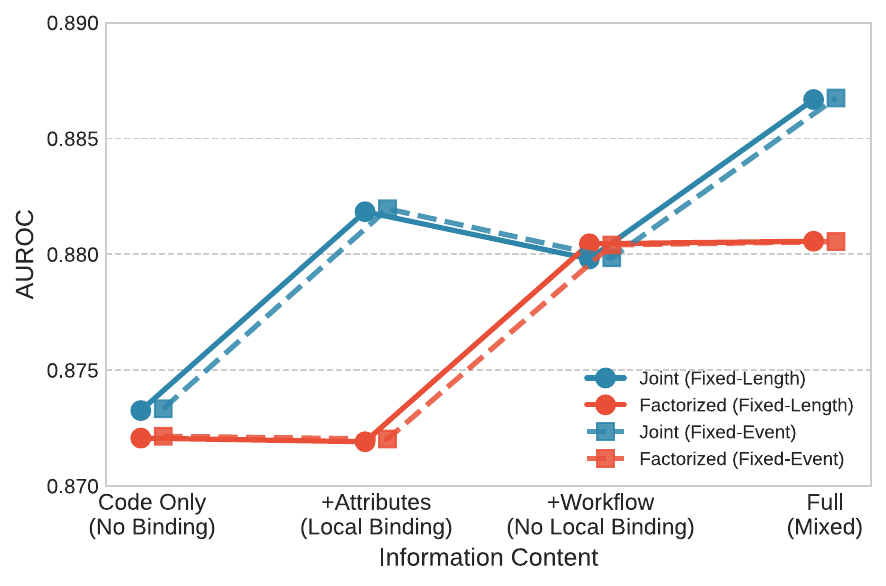}
\caption{Event encoding ablation. Mean AUROC for joint (blue) and factorized (red) event encoding as a function of information content (Code Only, +Attributes, +Workflow, Full), under fixed-length (solid line) and fixed-event (dashed line) sequence regimes. Time-Positions was used for time encoding. Error bars are omitted because they do not reflect variability in between-condition contrasts. Abbreviation: AUROC -- area under the receiver operating characteristic curve.}
\label{fig:fig3}
\end{figure}

A targeted ablation disentangled the contributions of token efficiency and local binding efficiency (Figure~\ref{fig:fig3}; Supplementary Table~S12). Performance was virtually identical between the Fixed-Length and Fixed-Event sequence regimes ($\beta = 0.000$, $P = 0.935$), indicating that differential truncation of patient history did not account for the encoding effect. At baseline (Code Only), joint and factorized encoding performed similarly ($\beta = -0.001$, $P = 0.230$). Adding value attributes improved joint encoding ($\beta = 0.009$, $P < 0.001$) but not factorized encoding (Factorized $\times$ Attributes interaction: $\beta = -0.009$, $P < 0.001$). In contrast, adding workflow stages improved both encoding strategies similarly (Factorized $\times$ Workflow interaction: $\beta = 0.002$, $P = 0.188$).

\subsection{Explicit Temporal Encoding Provides Modest Benefit Beyond Sequence Order}
\begin{figure}[H]
\centering
\includegraphics[width=\textwidth]{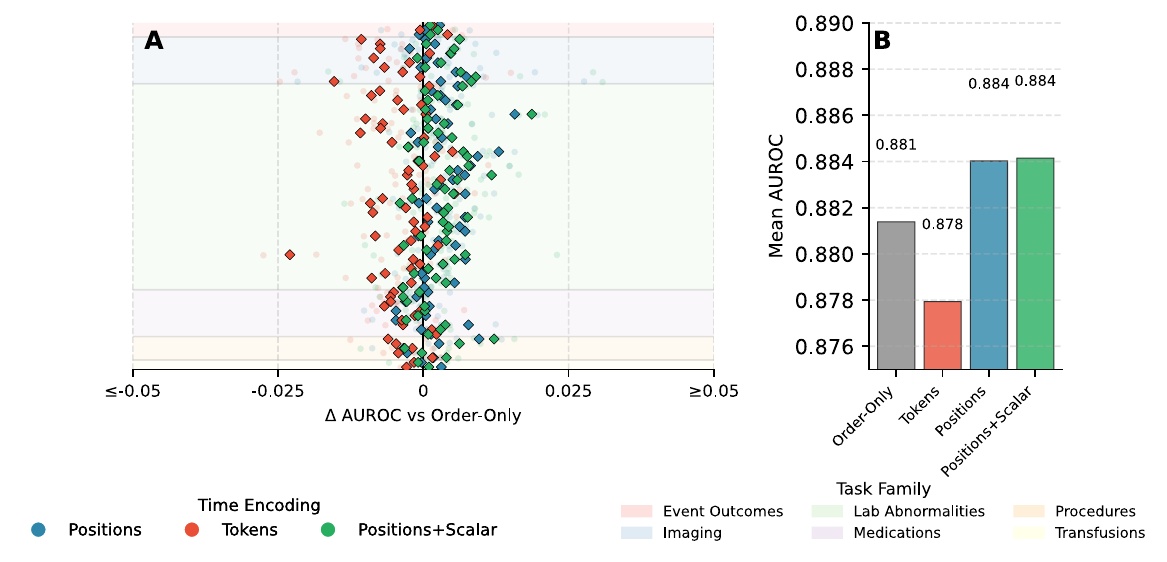}
\caption{Time encoding ablation. (A) Task-specific AUROC differences relative to Order-Only for three explicit time encoding strategies. Joint event encoding was used for each condition. Each transparent point represents the AUROC difference for individual evaluations. Opaque points denote the mean AUROC difference. Diamond color indicates time encoding strategy. Background bands indicate task family. (B) Mean AUROC by time encoding strategy. Error bars are omitted in the bar graph because they do not reflect variability in between-encoding contrasts. Abbreviations: AUROC -- area under the receiver operating characteristic curve.}
\label{fig:fig4}
\end{figure}

Relative to Order-Only encoding, which captures sequential ordering without temporal spacing, Time-Positions and Positions+Scalar each improved performance modestly and by nearly identical magnitudes ($\beta = 0.003$, $P < 0.001$ for both; Figure~\ref{fig:fig4}; Supplementary Table~S13). Time-Tokens performed significantly worse than Order-Only ($\beta = -0.003$, $P < 0.001$). Task-level effects were small across all strategies, with most tasks clustering near zero difference relative to Order-Only.

\subsection{Event Encoding Effects Generalize Across Institutions}
\begin{figure}[H]
\centering
\includegraphics[width=\textwidth]{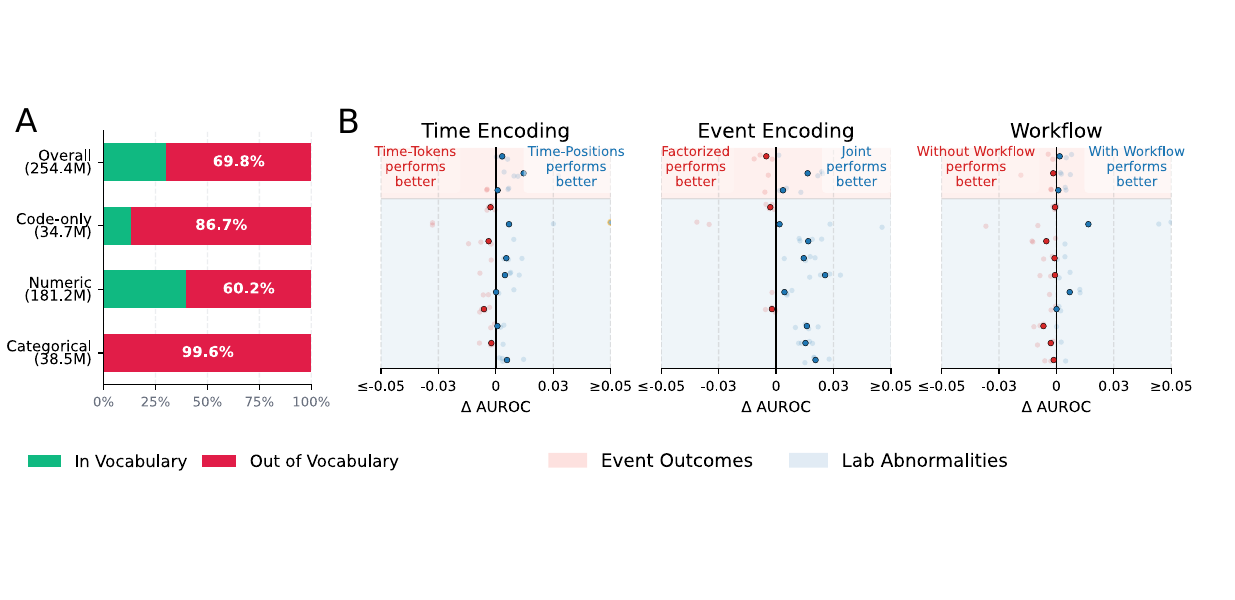}
\caption{External evaluation of SickKids foundation model on MIMIC. (A) Out-of-vocabulary (OOV) rates when applying tokenizers learned during SickKids pretraining to the full MIMIC dataset, stratified by event type. (B) AUROC differences between tokenization strategies for frozen SickKids-pretrained models evaluated on 13 MIMIC clinical prediction tasks. Each transparent point represents the AUROC difference for a single task under a specific experimental configuration. Opaque points denote the mean AUROC difference for each task, averaged across all other experimental factors. Background bands indicate task family. Absolute AUROCs by tokenization condition and task are reported in Supplementary Tables S14 and S15, respectively.}
\label{fig:fig5}
\end{figure}

External evaluation of SickKids-pretrained models on MIMIC revealed substantial vocabulary mismatch, with overall out-of-vocabulary rates of 69.8\% and exceeding 85\% for code-only and categorical text events (Figure~\ref{fig:fig5}A). The joint encoding advantage transferred with a similar effect size to the SickKids local evaluation setting ($\beta = 0.008$, 95\% CI [0.005, 0.011], $P < 0.001$, Figure~5B; Supplementary Tables~S14--S16), while time encoding ($\beta = 0.002$, $P = 0.271$) and workflow ($\beta < 0.001$, $P = 0.803$) effects were not significant. Frozen transfer achieved best mean AUROC of 0.815 (joint event encoding with Time-Positions and no workflow stages) compared to 0.842 (joint event encoding with Time-Positions) for MIMIC models trained from scratch with domain-matched vocabularies, all without workflow stages (Supplementary Tables~S17--S18).

\section{Discussion}
This study systematically evaluated how three tokenization design choices affect structured EHR foundation model performance across 74 clinical prediction tasks. Joint event encoding and Time-Positions outperformed their alternatives while requiring less pretraining compute. Workflow annotations improved local performance but did not transfer across institutions. Targeted ablations traced the joint encoding advantage to local binding efficiency, whereby code-attribute pairs are pre-computed as single tokens during tokenization rather than learned across tokens from limited training data. External evaluations demonstrated that this advantage generalized across institutions despite substantial vocabulary mismatch, while temporal and workflow effects did not. These results indicate that tokenization can meaningfully affect both the predictive performance and computational cost of structured EHR foundation models.

The better-performing strategies for event and time encoding were also more computationally efficient due to producing shorter token sequences. This suggests that tokens capturing meaningful domain structure are both more compact and learnable in data-limited settings. Recent research on language models has found that tokenizers that align boundaries with linguistically meaningful units (words, morphemes, common multi-word expressions) outperform those optimizing compression alone.\cite{goldman2024unpacking,schmidt2024tokenization,liu2025superbpe} Here, joint encoding fuses clinical codes and their measured values into tokens that correspond to meaningful clinical concepts, while Time-Positions encodes temporal information through the positional mechanism rather than consuming additional tokens. For both event and time encoding, the less effective tokenization strategy expands sequence length with tokens the model integrates less efficiently, increasing compute cost without proportional representational benefit. These efficiency differences are practically significant given tokenization choices are fixed at pretraining time and propagate to every downstream application.

The event encoding ablation revealed a specific dissociation: adding value attributes improved joint encoding but not factorized encoding, whereas adding workflow stages improved both strategies similarly. This pattern is consistent with a local binding problem applied to clinical event sequences.\cite{greff2020binding} Under factorized encoding, a shared attribute vocabulary is reused across thousands of clinical codes, requiring the same token to carry different clinical meaning depending on which code it follows. For instance, an elevated quantile for serum glucose indicates hyperglycemia, whereas the same quantile for creatinine indicates renal impairment. The model must learn these context-dependent associations from pretraining using next-token prediction. Conversely, workflow annotations arrive as separate events at distinct timestamps and may be independently informative, without requiring local compositional binding, which may explain why both encoding strategies benefited similarly. Joint encoding resolves the local binding problem by fusing the clinical code and its measured value into a single token, pre-computing their association at tokenization time. While transformers can in principle learn such local bindings, compositional generalization might require larger models trained on correspondingly larger datasets. Structured EHR data, which is orders of magnitude smaller than language corpora,\cite{wornow2023shaky,guo2026systematic} may be insufficient for models to reliably discover these associations, making it advantageous to resolve them at tokenization time.

Explicit temporal encoding provided modest benefit beyond sequence order, and discrete time tokens degraded performance. The latter is consistent with recent evaluations showing that explicit temporal encoding provides little benefit across supervised clinical prediction and EHR foundation model settings.\cite{alattrach2025rethinking,wornow2024contextclues} In our study, discrete tokens may degrade performance by fragmenting clinical event sequences and consuming context positions without proportional information gain. In contrast, Time-Positions provided a small improvement without additional tokens. However, this advantage did not transfer from the pediatric pretraining population to an adult ICU cohort, likely because age-calibrated positional encodings embed the temporal statistics of the pretraining population and become unreliable under extrapolation. These results suggest that how temporal information is encoded matters more than whether it is included, and that developing temporal representations robust to demographic shift remains an open challenge.

Workflow stages improved local performance but required 36\% more pretraining FLOPs due to the additional tokens and provided no benefit in transfer. The latter is likely because the temporal staging of orders, collections, and results varies across institutions and clinical workflows. Including workflow stages during pretraining did not degrade transfer performance when applied to data without workflow annotations, suggesting the model does not become dependent on this information. Whether mismatched workflow patterns between institutions could negatively affect transfer remains an open question.

During external evaluation, vocabulary coverage was poor, with nearly 70\% of MIMIC clinical codes out-of-vocabulary. Despite this, the joint event encoding advantage generalized with a comparable effect size, consistent with local binding efficiency being a generalizable benefit in modeling of structured clinical event sequences. Temporal and workflow effects were not significant, consistent with their dependence on population-specific or institution-specific characteristics.

These results suggest that vocabulary alignment should be prioritized for cross-institutional deployment and argue for explicit reporting of tokenization strategies in EHR foundation model studies.\cite{guo2026systematic} The consistent advantages of joint event encoding and positional time encoding across tasks suggest these as reasonable defaults for future model development. More broadly, the observation that pre-computed structural associations transfer more robustly than learned temporal or contextual patterns may extend beyond EHR to other domains where foundation models are applied to sequences of structured events with limited training data.

Several limitations warrant consideration. All models in the main experiments were pretrained on data from a single pediatric hospital. Evaluation was restricted to discriminative prediction tasks using linear probes on frozen representations and we did not evaluate generative applications such as zero-shot prediction.\cite{renc2024zeroshot,waxler2025generative,renc2025adaptive} However, because tokenization operates upstream of the pretraining objective and determines the information available to the model, the binding efficiency and temporal encoding findings are likely to hold regardless of downstream application. The external evaluation experiments confound vocabulary mismatch, population shift, and institutional differences, which cannot be fully disentangled. We did not investigate data-driven tokenization approaches such as byte-pair encoding. Recent work has begun adapting BPE to structured medical codes by merging frequently co-occurring codes into single tokens,\cite{dwivedi2024structured} but the interaction between such learned tokenization and the design axes evaluated here remains unexplored. Finally, while effect sizes were consistent across tasks, clinical significance at the individual task level should be interpreted cautiously.

In conclusion, our results demonstrate that encoding clinically meaningful structure at tokenization time consistently improves both performance and computational efficiency across 74 clinical prediction tasks. Pre-computing code-attribute associations into joint tokens avoids the local binding problem that factorized representations impose in data-limited settings, and this advantage transfers across institutions even under substantial vocabulary mismatch. As structured EHR foundation models advance toward clinical deployment, principled tokenization offers a tractable lever for improving robustness and reducing pretraining cost.

\section*{Data Availability}
The SickKids dataset cannot be made publicly available due to patient privacy restrictions. Relevant data are available upon reasonable request to the corresponding author. The MIMIC-IV dataset is publicly available through PhysioNet (\url{https://physionet.org/content/mimiciv/1.0/}) subject to credentialing and a data use agreement.

\section*{Code Availability}
The codebase for EHR tokenization and foundation model training will be made publicly available at \url{https://github.com/sungresearch/ehr-fm}.

\section*{Acknowledgements}
LS is supported by the Canada Research Chair in Pediatric Oncology Supportive Care.

\section*{Funding}
This research did not receive funding.

\section*{Author Contribution}
L.L.G. conceptualized and designed the study with input from all authors. L.L.G. performed all experiments, analyzed and interpreted results with input from all authors. L.L.G. wrote the manuscript with input from all authors. All authors read and approved the final manuscript.

\section*{Competing Interests}
The authors declare no competing interests.

\bibliographystyle{unsrt}
\bibliography{references}

@misc{bommasani2021,
  author = {Bommasani, Rishi and Hudson, Drew A. and Adeli, Ehsan and Altman, Russ and Arora, Simran and von Arx, Sydney and Bernstein, Michael S. and Bohg, Jeannette and Bosselut, Antoine and Brunskill, Emma and Brynjolfsson, Erik and Buch, Shyamal and Card, Dallas and Castellon, Rodrigo and Chatterji, Niladri S. and Chen, Annie S. and Creel, Kathleen A. and Davis, Jared and Demszky, Dora and Donahue, Chris and Doumbouya, Moussa and Durmus, Esin and Ermon, Stefano and Etchemendy, John and Ethayarajh, Kawin and Fei-Fei, Li and Finn, Chelsea and Gale, Trevor and Gillespie, Lauren and Goel, Karan and Goodman, Noah D. and Grossman, Shelby and Guha, Neel and Hashimoto, Tatsunori and Henderson, Peter and Hewitt, John and Ho, Daniel E. and Hong, Jenny and Hsu, Kyle and Huang, Jing and Icard, Thomas F. and Jain, Saahil and Jurafsky, Dan and Kalluri, Pratyusha and Karamcheti, Siddharth and Keeling, Geoff and Khani, Fereshte and Khattab, Omar and Koh, Pang Wei and Krass, Mark S. and Krishna, Ranjay and Kuditipudi, Rohith and Kumar, Ananya and Ladhak, Faisal and Lee, Mina and Lee, Tony and Leskovec, Jure and Levent, Isabelle and Li, Xiang Lisa and Li, Xuechen and Ma, Tengyu and Malik, Ali and Manning, Christopher D. and Mirchandani, Suvir and Mitchell, Eric and Munyikwa, Zanele and Nair, Suraj and Narayan, Avanika and Narayanan, Deepak and Newman, Ben and Nie, Allen and Niebles, Juan Carlos and Nilforoshan, Hamed and Nyarko, Julian and Ogut, Giray and Orr, Laurel J. and Papadimitriou, Isabel and Park, Joon Sung and Piech, Chris and Portelance, Eva and Potts, Christopher and Raghunathan, Aditi and Reich, Rob and Ren, Hongyu and Rong, Frieda and Roohani, Yusuf H. and Ruiz, Camilo and Ryan, Jack and R{\'e}, Christopher and Sadigh, Dorsa and Sagawa, Shiori and Santhanam, Keshav and Shih, Andy and Srinivasan, Krishnan P. and Tamkin, Alex and Taori, Rohan and Thomas, Armin W. and Tram{\`e}r, Florian and Wang, Rose E. and Wang, William and Wu, Bohan and Wu, Jiajun and Wu, Yuhuai and Xie, Sang Michael and Yasunaga, Michihiro and You, Jiaxuan and Zaharia, Matei A. and Zhang, Michael and Zhang, Tianyi and Zhang, Xikun and Zhang, Yuhui and Zheng, Lucia and Zhou, Kaitlyn and Liang, Percy},
  title = {On the Opportunities and Risks of Foundation Models},
  year = {2021},
  eprint = {2108.07258},
  archivePrefix = {arXiv},
  doi = {10.48550/arXiv.2108.07258},
  url = {https://doi.org/10.48550/arXiv.2108.07258}
}

@article{moor2023,
  author = {Moor, Michael and Banerjee, Oishi and Abad, Zahra Shakeri Hossein and Krumholz, Harlan M. and Leskovec, Jure and Topol, Eric and Rajpurkar, Pranav},
  title = {Foundation models for generalist medical artificial intelligence},
  journal = {Nature},
  volume = {616},
  pages = {259--265},
  year = {2023},
  doi = {10.1038/s41586-023-05881-4}
}

@article{wu2025,
  author = {Wu, Chenyu and Zhang, Xiaoman and Zhang, Ya and Hui, Hui and Wang, Yanfeng and Xie, Weidi},
  title = {Towards generalist foundation model for radiology by leveraging web-scale 2D\&3D medical data},
  journal = {Nature Communications},
  volume = {16},
  pages = {7866},
  year = {2025},
  doi = {10.1038/s41467-025-62385-7}
}

@article{fu2025,
  author = {Fu, Xi and Mo, Shentong and Buendia, Alejandro and Laurent, Anouchka P. and Shao, Anqi and Alvarez-Torres, Maria del Mar and Yu, Tianji and Tan, Jimin and Su, Jiayu and Sagatelian, Romella and Ferrando, Adolfo A. and Ciccia, Alberto and Lan, Yanyan and Owens, David M. and Palomero, Teresa and Xing, Eric P. and Rabadan, Raul},
  title = {A foundation model of transcription across human cell types},
  journal = {Nature},
  volume = {637},
  pages = {965--973},
  year = {2025},
  doi = {10.1038/s41586-024-08391-z}
}

@article{singhal2025,
  author = {Singhal, Karan and others},
  title = {Toward expert-level medical question answering with large language models},
  journal = {Nature Medicine},
  volume = {31},
  pages = {943--950},
  year = {2025},
  doi = {10.1038/s41591-024-03423-7}
}

@article{steinberg2021,
  author = {Steinberg, Ethan and Jung, Ken and Fries, Jason A. and Corbin, Conor K. and Pfohl, Stephen R. and Shah, Nigam H.},
  title = {Language models are an effective representation learning technique for electronic health record data},
  journal = {Journal of Biomedical Informatics},
  volume = {113},
  pages = {103637},
  year = {2021},
  doi = {10.1016/j.jbi.2020.103637}
}

@misc{steinberg2023motor,
  author = {Steinberg, Ethan and Fries, Jason and Xu, Yizhe and Shah, Nigam},
  title = {MOTOR: A Time-To-Event Foundation Model For Structured Medical Records},
  year = {2023},
  eprint = {2301.03150},
  archivePrefix = {arXiv},
  doi = {10.48550/arXiv.2301.03150},
  url = {https://doi.org/10.48550/arXiv.2301.03150}
}

@article{guo2023shift,
  author = {Guo, Lin Lawrence and Steinberg, Ethan and Fleming, Scott L. and Posada, Jose and Lemmon, Joshua and Pfohl, Stephen R. and Shah, Nigam H. and Fries, Jason and Sung, Lillian},
  title = {EHR foundation models improve robustness in the presence of temporal distribution shift},
  journal = {Scientific Reports},
  volume = {13},
  pages = {3767},
  year = {2023},
  doi = {10.1038/s41598-023-30820-8}
}

@article{guo2024multicenter,
  author = {Guo, Lin Lawrence and Fries, Jason A. and Steinberg, Ethan and Fleming, Scott L. and Morse, Keith and Aftandilian, Catherine and Posada, Jose and Shah, Nigam H. and Sung, Lillian},
  title = {A multi-center study on the adaptability of a shared foundation model for electronic health records},
  journal = {npj Digital Medicine},
  volume = {7},
  pages = {171},
  year = {2024},
  doi = {10.1038/s41746-024-01166-w}
}

@article{lemmon2023selfsupervised,
  author = {Lemmon, Joshua and Guo, Lin Lawrence and Steinberg, Ethan and Morse, Keith E. and Fleming, Scott Lanyon and Aftandilian, Catherine and Pfohl, Stephen R. and Posada, Jose D. and Shah, Nigam and Fries, Jason and Sung, Lillian},
  title = {Self-supervised machine learning using adult inpatient data produces effective models for pediatric clinical prediction tasks},
  journal = {Journal of the American Medical Informatics Association},
  volume = {30},
  pages = {2004--2011},
  year = {2023},
  doi = {10.1093/jamia/ocad175}
}

@inproceedings{wornow2023neurips,
  author = {Wornow, Michael and Thapa, Rahul and Steinberg, Ethan and Fries, Jason A. and Shah, Nigam H.},
  title = {EHRSHOT: An EHR Benchmark for Few-Shot Evaluation of Foundation Models},
  booktitle = {Proceedings of the 37th International Conference on Neural Information Processing Systems},
  year = {2023}
}

@article{wornow2023shaky,
  author = {Wornow, Michael and Xu, Yizhe and Thapa, Rahul and Patel, Birju and Steinberg, Ethan and Fleming, Scott and Pfeffer, Michael A. and Fries, Jason and Shah, Nigam H.},
  title = {The shaky foundations of large language models and foundation models for electronic health records},
  journal = {npj Digital Medicine},
  volume = {6},
  pages = {135},
  year = {2023},
  doi = {10.1038/s41746-023-00879-8}
}

@article{kim2025pretrained,
  author = {Kim, Junmo and Kim, Joo Seong and Lee, Ji-Hyang and Kim, Min-Gyu and Kim, Taehyun and Cho, Chaeeun and Park, Rae Woong and Kim, Kwangsoo},
  title = {Pretrained patient trajectories for adverse drug event prediction using common data model-based electronic health records},
  journal = {Communications Medicine},
  volume = {5},
  pages = {232},
  year = {2025},
  doi = {10.1038/s43856-025-00914-7}
}

@article{renc2024zeroshot,
  author = {Renc, Pawel and Jia, Yugang and Samir, Anthony E. and Was, Jaroslaw and Li, Quanzheng and Bates, David W. and Sitek, Arkadiusz},
  title = {Zero shot health trajectory prediction using transformer},
  journal = {npj Digital Medicine},
  volume = {7},
  pages = {256},
  year = {2024},
  doi = {10.1038/s41746-024-01235-0}
}

@misc{waxler2025generative,
  author = {Waxler, Shane and Blazek, Paul and White, Davis and Sneider, Daniel and Chung, Kevin and Nagarathnam, Mani and Williams, Patrick and Voeller, Hank and Wong, Karen and Swanhorst, Matthew and Zhang, Sheng and Usuyama, Naoto and Wong, Cliff and Naumann, Tristan and Poon, Hoifung and Loza, Andrew and Meeker, Daniella and Hain, Seth and Shah, Rahul},
  title = {Generative Medical Event Models Improve with Scale},
  year = {2025},
  eprint = {2508.12104},
  archivePrefix = {arXiv}
}

@article{hur2024genhpf,
  author = {Hur, Kyunghoon and Oh, Jungwoo and Kim, Junu and Kim, Jiyoun and Lee, Min Jae and Cho, Eunbyeol and Moon, Seong-Eun and Kim, Young-Hak and Atallah, Louis and Choi, Edward},
  title = {GenHPF: General Healthcare Predictive Framework for Multi-Task Multi-Source Learning},
  journal = {IEEE Journal of Biomedical and Health Informatics},
  volume = {28},
  pages = {502--513},
  year = {2024},
  doi = {10.1109/JBHI.2023.3327951}
}

@misc{alattrach2025rethinking,
  author = {Al Attrach, Rafi and Fani, Rajna and Restrepo, David and Jia, Yugang and Sch\"uffler, Peter},
  title = {Rethinking Tokenization for Clinical Time Series: When Less is More},
  year = {2025},
  eprint = {2512.05217},
  archivePrefix = {arXiv},
  doi = {10.48550/arXiv.2512.05217},
  url = {https://doi.org/10.48550/arXiv.2512.05217}
}

@misc{pang2025cehrxgpt,
  author = {Pang, Chao and Park, Jiheum and Jiang, Xinzhou and Pavinkurve, Nishanth Parameshwar and Kalluri, Krishna S. and Joshi, Shalmali and Elhadad, Noemie and Natarajan, Karthik},
  title = {CEHR-XGPT: A Scalable Multi-Task Foundation Model for Electronic Health Records},
  year = {2025},
  eprint = {2509.03643},
  archivePrefix = {arXiv}
}

@inproceedings{pang2021cehrbert,
  author = {Pang, Chao and Jiang, Xinzhuo and Kalluri, Krishna S. and Spotnitz, Matthew and Chen, RuiJun and Perotte, Adler and Natarajan, Karthik},
  title = {CEHR-BERT: Incorporating temporal information from structured EHR data to improve prediction tasks},
  booktitle = {Proceedings of Machine Learning for Health},
  volume = {158},
  series = {Proceedings of Machine Learning Research},
  pages = {239--260},
  year = {2021},
  publisher = {PMLR},
  url = {https://proceedings.mlr.press/v158/pang21a.html}
}

@misc{wornow2024contextclues,
  author = {Wornow, Michael and Bedi, Suhana and Fuentes Hernandez, Miguel Angel and Steinberg, Ethan and Fries, Jason Alan and R{\'e}, Christopher and Koyejo, Sanmi and Shah, Nigam H.},
  title = {Context Clues: Evaluating Long Context Models for Clinical Prediction Tasks on EHRs},
  year = {2024},
  eprint = {2412.16178},
  archivePrefix = {arXiv}
}

@article{kraljevic2024foresight,
  author = {Kraljevic, Zeljko and Bean, Dan and Shek, Anthony and Bendayan, Rebecca and Hemingway, Harry and Au Yeung, Joshua and Deng, Alexander and Baston, Alfred and Ross, Jack and Idowu, Esther and Teo, James T. and Dobson, Richard J. B.},
  title = {Foresight---a generative pretrained transformer for modelling of patient timelines using electronic health records: a retrospective modelling study},
  journal = {Lancet Digital Health},
  volume = {6},
  pages = {e281--e290},
  year = {2024},
  doi = {10.1016/S2589-7500(24)00025-6}
}

@article{rasmy2021medbert,
  author = {Rasmy, Laila and Xiang, Yang and Xie, Ziqian and Tao, Cui and Zhi, Degui},
  title = {Med-BERT: pretrained contextualized embeddings on large-scale structured electronic health records for disease prediction},
  journal = {npj Digital Medicine},
  volume = {4},
  pages = {86},
  year = {2021},
  doi = {10.1038/s41746-021-00455-y}
}

@article{guo2026systematic,
  author = {Guo, Lin Lawrence and Arciniegas, Santiago Eduardo and Yan, Adam Paul and Fries, Jason A. and Tomlinson, George A. and Sung, Lillian},
  title = {Systematic Review of Foundation Models for Structured Electronic Health Records},
  journal = {Journal of the American Medical Informatics Association},
  year = {2026},
  doi = {https://doi.org/10.1093/jamia/ocag033}
}

@misc{greff2020binding,
  author = {Greff, Klaus and van Steenkiste, Sjoerd and Schmidhuber, J\"urgen},
  title = {On the Binding Problem in Artificial Neural Networks},
  year = {2020},
  eprint = {2012.05208},
  archivePrefix = {arXiv},
  doi = {10.48550/arXiv.2012.05208},
  url = {https://doi.org/10.48550/arXiv.2012.05208}
}

@article{johnson2023mimiciv,
  author = {Johnson, Alistair E. W. and Bulgarelli, Lucas and Shen, Lu and Gayles, Alvin and Shammout, Ayad and Horng, Steven and Pollard, Tom J. and Hao, Sicheng and Moody, Benjamin and Gow, Brian and Lehman, Li-wei H. and Celi, Leo A. and Mark, Roger G.},
  title = {MIMIC-IV, a freely accessible electronic health record dataset},
  journal = {Scientific Data},
  volume = {10},
  pages = {1},
  year = {2023},
  doi = {10.1038/s41597-022-01899-x}
}

@article{guo2023sedar,
  author = {Guo, Lin Lawrence and Calligan, Maryann and Vettese, Emily and Cook, Sadie and Gagnidze, George and Han, Oscar and Inoue, Jiro and Lemmon, Joshua and Li, Johnson and Roshdi, Medhat and Sadovy, Bohdan and Wallace, Steven and Sung, Lillian},
  title = {Development and validation of the SickKids Enterprise-wide Data in Azure Repository (SEDAR)},
  journal = {Heliyon},
  volume = {9},
  pages = {e21586},
  year = {2023},
  doi = {10.1016/j.heliyon.2023.e21586}
}

@inproceedings{arnrich2024meds,
  author = {Arnrich, Bert and Choi, Edward and Fries, Jason Alan and McDermott, Matthew B. B. and Oh, Jungwoo and Pollard, Tom and Shah, Nigam and Steinberg, Ethan and Wornow, Michael and van de Water, Robin},
  title = {Medical Event Data Standard (MEDS): Facilitating Machine Learning for Health},
  booktitle = {ICLR 2024 Workshop on Learning from Time Series for Health},
  year = {2024}
}

@misc{steinberg2024medsreader,
  author = {Steinberg, Ethan and Wornow, Michael and Bedi, Suhana and Fries, Jason A. and McDermott, Matthew and Shah, Nigam H.},
  title = {meds\_reader: A fast and efficient EHR processing library},
  year = {2024},
  eprint = {2409.09095},
  archivePrefix = {arXiv},
  doi = {10.48550/arXiv.2409.09095},
  url = {https://doi.org/10.48550/arXiv.2409.09095}
}

@misc{ohdsi2021mimic,
  author = {OHDSI},
  title = {MIMIC},
  year = {2021},
  url = {https://github.com/OHDSI/MIMIC}
}

@article{goldberger2000physiobank,
  author = {Goldberger, Ary L. and Amaral, Luis A. N. and Glass, Leon and Hausdorff, Jeffrey M. and Ivanov, Plamen Ch. and Mark, Roger G. and Mietus, Joseph E. and Moody, George B. and Peng, Chung-Kang and Stanley, H. Eugene},
  title = {PhysioBank, PhysioToolkit, and PhysioNet: components of a new research resource for complex physiologic signals},
  journal = {Circulation},
  volume = {101},
  number = {23},
  pages = {E215--E220},
  year = {2000},
  doi = {10.1161/01.CIR.101.23.E215}
}

@article{yan2025roadmap,
  author = {Yan, Adam P. and Guo, Lin Lawrence and Inoue, Jiro and Arciniegas, Santiago E. and Vettese, Emily and Wolochacz, Agata and Crellin-Parsons, Nicole and Purves, Brandon and Wallace, Steven and Patel, Azaz and Roshdi, Medhat and Jessa, Karim and Cardiff, Bren and Sung, Lillian},
  title = {A roadmap to implementing machine learning in healthcare: from concept to practice},
  journal = {Frontiers in Digital Health},
  volume = {7},
  pages = {1462751},
  year = {2025},
  doi = {10.3389/fdgth.2025.1462751}
}

@article{su2024roformer,
  author = {Su, Jianlin and Ahmed, Murtadha and Lu, Yu and Pan, Shengfeng and Bo, Wen and Liu, Yunfeng},
  title = {RoFormer: Enhanced Transformer with Rotary Position Embedding},
  journal = {Neurocomputing},
  volume = {568},
  pages = {127063},
  year = {2024},
  doi = {10.1016/j.neucom.2023.127063}
}

@inproceedings{vaswani2017attention,
  author = {Vaswani, Ashish and Shazeer, Noam and Parmar, Niki and Uszkoreit, Jakob and Jones, Llion and Gomez, Aidan N. and Kaiser, Lukasz and Polosukhin, Illia},
  title = {Attention Is All You Need},
  booktitle = {Advances in Neural Information Processing Systems},
  volume = {30},
  year = {2017}
}

@inproceedings{brown2020language,
  author = {Brown, Tom B. and Mann, Benjamin and Ryder, Nick and Subbiah, Melanie and Kaplan, Jared and Dhariwal, Prafulla and Neelakantan, Arvind and Shyam, Pranav and Sastry, Girish and Askell, Amanda and Agarwal, Sandhini and Herbert-Voss, Ariel and Krueger, Gretchen and Henighan, Tom and Child, Rewon and Ramesh, Aditya and Ziegler, Daniel M. and Wu, Jeffrey and Winter, Clemens and Hesse, Christopher and Chen, Mark and Sigler, Eric and Litwin, Mateusz and Gray, Scott and Chess, Benjamin and Clark, Jack and Berner, Christopher and McCandlish, Sam and Radford, Alec and Sutskever, Ilya and Amodei, Dario},
  title = {Language Models are Few-Shot Learners},
  booktitle = {Advances in Neural Information Processing Systems},
  volume = {33},
  pages = {1877--1901},
  year = {2020}
}

@inproceedings{warner2024modernbert,
  author = {Warner, Benjamin and Chaffin, Antoine and Clavi\'e, Benjamin and Weller, Orion and Hallstr\"om, Oskar and Taghadouini, Said and Gallagher, Alexis and Biswas, Raja and Ladhak, Faisal and Aarsen, Tom and Cooper, Nathan and Adams, Griffin and Howard, Jeremy and Poli, Iacopo},
  title = {Smarter, Better, Faster, Longer: A Modern Bidirectional Encoder for Fast, Memory Efficient, and Long Context Finetuning and Inference},
  booktitle = {Proceedings of the 63rd Annual Meeting of the Association for Computational Linguistics},
  pages = {2526--2547},
  year = {2025},
  publisher = {Association for Computational Linguistics},
  eprint = {2412.13663},
  archivePrefix = {arXiv}
}

@misc{child2019sparse,
  author = {Child, Rewon and Gray, Scott and Radford, Alec and Sutskever, Ilya},
  title = {Generating Long Sequences with Sparse Transformers},
  year = {2019},
  eprint = {1904.10509},
  archivePrefix = {arXiv}
}

@inproceedings{muennighoff2023scaling,
  author = {Muennighoff, Niklas and Rush, Alexander M. and Barak, Boaz and Le Scao, Teven and Piktus, Aleksandra and Tazi, Nouamane and Pyysalo, Sampo and Wolf, Thomas and Raffel, Colin},
  title = {Scaling Data-Constrained Language Models},
  booktitle = {Advances in Neural Information Processing Systems},
  volume = {36},
  pages = {50358--50376},
  year = {2023}
}

@misc{kaplan2020scaling,
  author = {Kaplan, Jared and McCandlish, Sam and Henighan, Tom and Brown, Tom B. and Chess, Benjamin and Child, Rewon and Gray, Scott and Radford, Alec and Wu, Jeffrey and Amodei, Dario},
  title = {Scaling Laws for Neural Language Models},
  year = {2020},
  eprint = {2001.08361},
  archivePrefix = {arXiv}
}

@inproceedings{paszke2019pytorch,
  author = {Paszke, Adam and Gross, Sam and Massa, Francisco and Lerer, Adam and Bradbury, James and Chanan, Gregory and Killeen, Trevor and Lin, Zeming and Gimelshein, Natalia and Antiga, Luca and Desmaison, Alban and Kopf, Andreas and Yang, Edward and DeVito, Zachary and Raison, Martin and Tejani, Alykhan and Chilamkurthy, Sasank and Steiner, Benoit and Fang, Lu and Bai, Junjie and Chintala, Soumith},
  title = {PyTorch: An Imperative Style, High-Performance Deep Learning Library},
  booktitle = {Advances in Neural Information Processing Systems},
  volume = {32},
  pages = {8024--8035},
  year = {2019}
}

@article{pedregosa2011scikitlearn,
  author = {Pedregosa, Fabian and Varoquaux, Ga\"el and Gramfort, Alexandre and Michel, Vincent and Thirion, Bertrand and Grisel, Olivier and Blondel, Mathieu and Prettenhofer, Peter and Weiss, Ron and Dubourg, Vincent and Vanderplas, Jake and Passos, Alexandre and Cournapeau, David and Brucher, Matthieu and Perrot, Matthieu and Duchesnay, \'Edouard},
  title = {Scikit-learn: Machine Learning in Python},
  journal = {Journal of Machine Learning Research},
  volume = {12},
  pages = {2825--2830},
  year = {2011}
}

@inproceedings{goldman2024unpacking,
  author = {Goldman, Omer and Caciularu, Avi and Eyal, Matan and Cao, Kris and Szpektor, Idan and Tsarfaty, Reut},
  title = {Unpacking Tokenization: Evaluating Text Compression and its Correlation with Model Performance},
  booktitle = {Proceedings of the 62nd Annual Meeting of the Association for Computational Linguistics},
  year = {2024}
}

@inproceedings{schmidt2024tokenization,
  author = {Schmidt, Craig W. and Reddy, Varshini and Zhang, Haoran and Alameddine, Alec and Uzan, Omri and Pinter, Yuval and Tanner, Chris},
  title = {Tokenization Is More Than Compression},
  booktitle = {Proceedings of the 2024 Conference on Empirical Methods in Natural Language Processing},
  year = {2024}
}

@misc{liu2025superbpe,
  author = {Liu, Alisa and Hayase, Jonathan and Hofmann, Valentin and Oh, Sewoong and Smith, Noah A. and Choi, Yejin},
  title = {SuperBPE: Space Travel for Language Models},
  year = {2025},
  eprint = {2503.13423},
  archivePrefix = {arXiv}
}

@article{renc2025adaptive,
  author = {Renc, Pawel and Grzeszczyk, Michal K and Oufattole, Nassim and Goode, Deirdre and Jia, Yugang and Bieganski, Szymon and McDermott, Matthew B.A. and Was, Jaroslaw and Samir, Anthony E. and Cunningham, Jonathan W. and Bates, David W. and Sitek, Arkadiusz},
  title = {Foundation model of electronic medical records for adaptive risk estimation},
  journal = {GigaScience},
  volume = {14},
  year = {2025},
  doi = {10.1093/gigascience/giaf107}
}

@misc{dwivedi2024structured,
  author = {Dwivedi, Vijay Prakash and Schlegel, Viktor and Liu, Andy T. and Nguyen, Thanh-Tung and Kashyap, Abhinav Ramesh and Wei, Jeng and Yin, Wei-Hsian and Winkler, Stefan and Tan, Robby T.},
  title = {Representation Learning of Structured Data for Medical Foundation Models},
  year = {2024},
  eprint = {2410.13351},
  archivePrefix = {arXiv}
}

\clearpage
\appendix
\section*{Supplementary Material}

\setcounter{figure}{0}
\renewcommand{\thefigure}{S\arabic{figure}}
\setcounter{table}{0}
\renewcommand{\thetable}{S\arabic{table}}

\subsection*{Supplementary Figure S1. Cohort Construction for Pretraining and Downstream Evaluation}
\begin{figure}[H]
\centering
\IfFileExists{S_Consort_Diagram.pdf}{%
  \includegraphics[width=\textwidth]{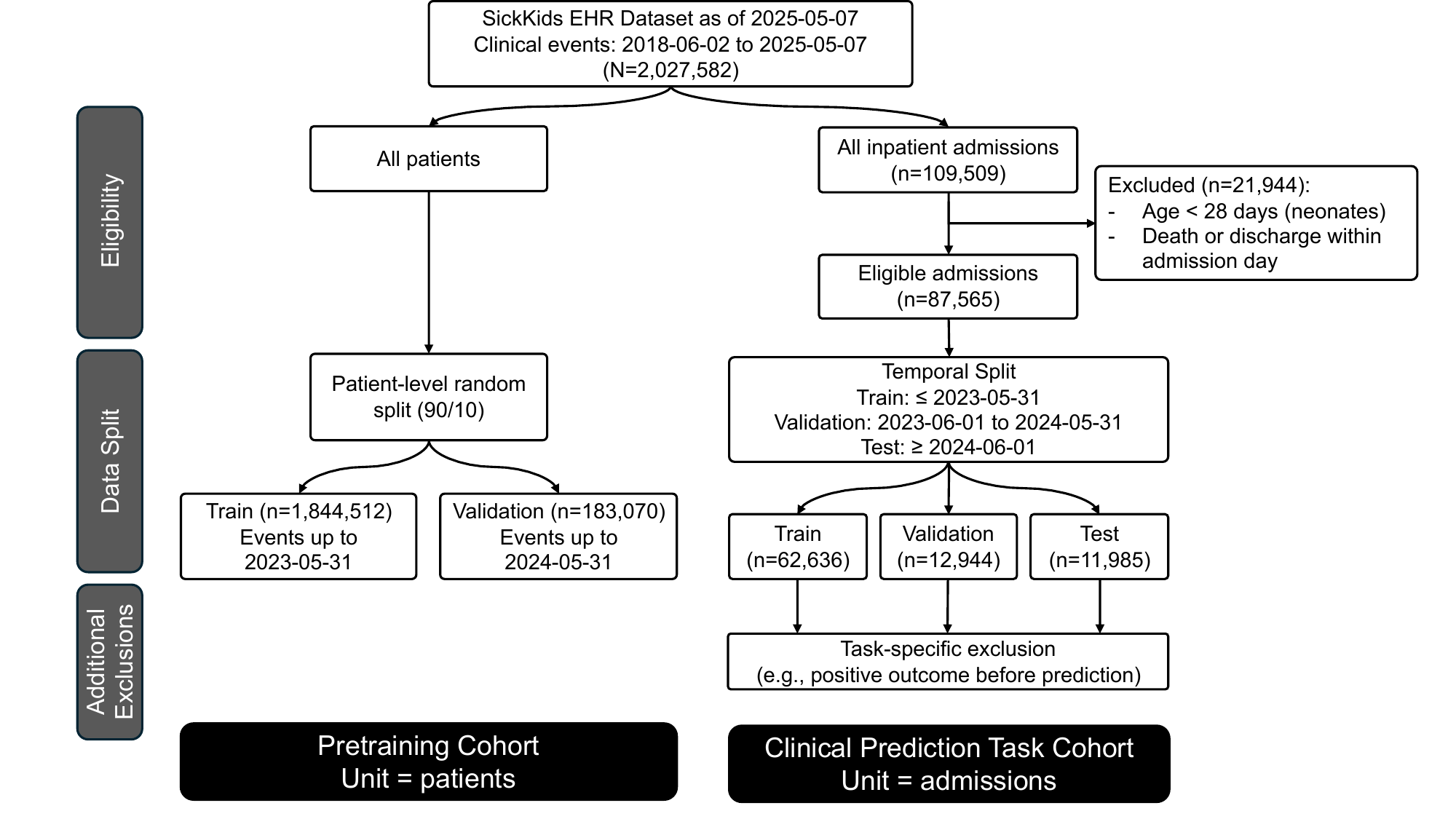}
}{%
  \fbox{\parbox{0.95\textwidth}{\centering Placeholder for Supplementary Figure S1\\
  Expected file: \texttt{SF1\_cohort\_construction.pdf} (not currently present in workspace).}}
}
\caption{Cohort Construction for Pretraining and Downstream Evaluation. Pretraining (left) used a patient-level random split, with different temporal cutoffs for training and validation to evaluate models on more recent clinical data during pretraining. Downstream evaluation (right) used admission-level temporal splits across 74 clinical prediction tasks, aligned to the same calendar cutoffs as pretraining. Task-specific exclusions (e.g., outcomes occurring before prediction time) were applied after cohort assignment. The unit of analysis was patients for pretraining and admissions for downstream prediction.}
\label{fig:supp_s1}
\end{figure}

\begin{table}[H]
\centering
\small
\caption*{Supplementary Table S1. Discrete Time-Interval Tokens Used in the Time-Tokens Conditions}
\begin{tabular}{ll}
\toprule
\textbf{Token Label*} & \textbf{Duration Range} \\
\midrule
INT\_5m\_15m** & 5 minutes to 15 minutes \\
INT\_15m\_1h & 15 minutes to 1 hour \\
INT\_1h\_2h & 1 hour to 2 hours \\
INT\_2h\_6h & 2 hours to 6 hours \\
INT\_6h\_12h & 6 hours to 12 hours \\
INT\_12h\_1d & 12 hours to 1 day \\
INT\_1d\_3d & 1 day to 3 days \\
INT\_3d\_1w & 3 days to 1 week \\
INT\_1w\_2w & 1 week to 2 weeks \\
INT\_2w\_1mt & 2 weeks to 1 month \\
INT\_1mt\_3mt & 1 month to 3 months \\
INT\_3mt\_6mt & 3 months to 6 months \\
INT\_6mt & 6 months (repeatable for longer durations) \\
\bottomrule
\end{tabular}
\par\vspace{0.4em}
{\footnotesize\raggedright\noindent  * Time-interval bins and token insertion rules were based on prior work \cite{renc2024zeroshot}. Specifically, no interval token was inserted when the elapsed time between events was shorter than 5 minutes. When the elapsed time exceeded 6 months, multiple 6-month interval tokens were inserted to approximate the total duration. Otherwise, a single interval token corresponding to the appropriate bin was used.}
\end{table}

\begin{table}[H]
\centering
\small
\caption*{Supplementary Table S2. SickKids Workflow-stage Event Representations by Clinical Domain}
\begin{tabular}{p{0.28\textwidth}p{0.28\textwidth}p{0.34\textwidth}}
\toprule
\textbf{Clinical Domain*} & \textbf{Without Workflow} & \textbf{With Workflow**} \\
\midrule
Measurement (Laboratory tests) & Result & Order $\rightarrow$ Taken $\rightarrow$ Result \\
Measurement (Flowsheets) & Single event & Single event \\
Procedure (Non-surgeries) & Start & Order $\rightarrow$ Start \\
Procedure (Surgeries) & Start & Start $\rightarrow$ End \\
Drug (Medication administrations) & Administration & Order $\rightarrow$ Administration \\
Drug (Prescriptions) & Start & Order $\rightarrow$ Start \\
Visit & Start, End & Start, End \\
Diagnosis & Single event & Single event \\
Observation & Single event & Single event \\
Visit Detail & Single event & Single event \\
\bottomrule
\end{tabular}
\par\vspace{0.4em}
{\footnotesize\noindent  * Clinical domains are grouped by OMOP domain, with sub-domain qualifiers where workflow patterns differ (e.g., laboratory tests vs. flowsheets). Domains without multi-step workflows are unchanged across conditions. \\ 
*** Arrows indicate the temporal ordering of workflow-stage events within a clinical action.}
\end{table}

\begin{table}[H]
\centering
\scriptsize
\caption*{Supplementary Table S3. Model Parameter Counts}
\begin{tabular}{lllrrrrr}
\toprule
Event & Time & Workflow & Embedding & Backbone* & NTP head & Total & Percent vocab.-dependent** \\
\midrule
Joint & Positions & No & 19.4M & 115.6M & 19.4M & 154.4M & 25.1\% \\
Joint & Positions & Yes & 27.1M & 115.6M & 27.2M & 169.9M & 32.0\% \\
Joint & Tokens & No & 19.4M & 115.6M & 19.4M & 154.5M & 25.2\% \\
Joint & Tokens & Yes & 27.2M & 115.6M & 27.2M & 170.0M & 32.0\% \\
Factorized & Positions & No & 8.36M & 115.6M & 8.37M & 132.4M & 12.6\% \\
Factorized & Positions & Yes & 8.99M & 115.6M & 9.01M & 133.6M & 13.5\% \\
Factorized & Tokens & No & 8.40M & 115.6M & 8.42M & 132.4M & 12.7\% \\
Factorized & Tokens & Yes & 9.04M & 115.6M & 9.06M & 133.7M & 13.5\% \\
\bottomrule
\end{tabular}
\par\vspace{0.4em}
{\footnotesize\noindent  * Transformer backbone parameters correspond to the transformer architecture and are held constant across all experiments. \\
 *** Vocabulary-dependent parameters include the embedding layer (input token embedding matrix) and the next-token prediction head during pretraining. Differences in total parameter count arise from changes in vocabulary size across tokenization strategies \\ 
 Abbreviations: NTP -- next-token prediction; M -- million.}
\end{table}

\begin{table}[H]
\centering
\small
\caption*{Supplementary Table S4. Training, Model, and Evaluation Hyperparameters}
\begin{tabular}{p{0.58\textwidth}p{0.32\textwidth}}
\toprule
\textbf{Hyperparameter} & \textbf{Value} \\
\midrule
\multicolumn{2}{l}{\textbf{Pretraining optimization and batching}} \\
Optimizer & AdamW \\
Learning rate & 5e-4 \\
Learning rate scheduler & cosine\_with\_min\_lr \\
Num epochs & 5 \\
Gradient accumulation steps & 2 \\
Early stopping & None \\
*Max tokens per batch & 32,768 \\
*Min patients per batch & 1 \\
Weight decay & 0.05 \\
Max gradient norm (clipping) & 1.0 \\
Warmup steps & 150 \\
Adam $\beta_1$ & 0.9 \\
Adam $\beta_2$ & 0.95 \\
Floating-point format & bf16 \\
\multicolumn{2}{l}{\textbf{Transformer backbone}} \\
Hidden size & 768 \\
Num layers & 28 \\
Num attention heads & 12 \\
Intermediate size & 1,152 \\
Activation & GELU \\
**Alternating dense layers & Yes \\
**Dense every n layers & 3 \\
**Attention width & 128 \\
\multicolumn{2}{l}{\textbf{Linear probe (Logistic Regression)}} \\
Input preprocessing & StandardScaler \\
Solver & LBFGS \\
Regularization & L2 \\
Inverse regularization (C) & 1, 0.1, 0.01, 0.001, 0.0001 \\
Max iterations & 10,000 \\
\bottomrule
\end{tabular}
\par\vspace{0.4em}
{\footnotesize\noindent  * Batches were constructed using a fixed token budget (32,768 tokens per batch) with a minimum of one patient per batch to accommodate variable-length patient sequences without padding. This imposes an effective upper bound of 32,768 tokens per patient sequence (i.e., maximum context window). When multiple patient sequences were in a batch, causal masking prevented attention across patient boundaries, ensuring independent sequence modeling.\\
*** The transformer alternates between global and local self-attention, starting with a global attention layer, followed by three local attention layers (attention width of 128), and repeating this pattern throughout.
}
\end{table}

\small
\begin{longtable}{lrrrr}
\caption*{Supplementary Table S5. SickKids and MIMIC Task Cohort Statistics}\\
\toprule
\textbf{Task} & \textbf{Total Admissions} & \textbf{Total Patients} & \textbf{Positive Cases} & \textbf{Prevalence (\%)} \\
\midrule
\endfirsthead
\toprule
\textbf{Task} & \textbf{Total Admissions} & \textbf{Total Patients} & \textbf{Positive Cases} & \textbf{Prevalence (\%)} \\
\midrule
\endhead
\multicolumn{5}{l}{\textbf{SickKids}}\\
\multicolumn{5}{l}{\textbf{Blood Bank}}\\
Platelet transfusion & 86030 & 50720 & 2932 & 3.41 \\
Red cell transfusion & 81077 & 48223 & 6487 & 8.00 \\
\multicolumn{5}{l}{\textbf{Procedure}}\\
Invasive intubation & 84294 & 49537 & 1557 & 1.85 \\
Gastrostomy tube & 87401 & 51177 & 425 & 0.49 \\
Echocardiogram & 85630 & 50296 & 6757 & 7.89 \\
Pulmonary function test & 87418 & 51178 & 774 & 0.89 \\
Lumbar puncture & 86577 & 50857 & 2233 & 2.58 \\
Surgery & 67918 & 39119 & 11958 & 17.61 \\
Interventional radiology & 86539 & 50989 & 7339 & 8.48 \\
\multicolumn{5}{l}{\textbf{Imaging}}\\
Plain radiography chest & 78058 & 46552 & 9129 & 11.70 \\
Ultrasound abdomen & 85025 & 50238 & 10640 & 12.51 \\
Computerized tomography chest & 87115 & 51084 & 2046 & 2.35 \\
Computerized tomography abdomen & 87103 & 51066 & 1233 & 1.42 \\
Computerized tomography head & 86174 & 50601 & 2900 & 3.37 \\
MRI head & 85777 & 50391 & 5237 & 6.11 \\
MRI whole body & 87441 & 51178 & 158 & 0.18 \\
PET & 87444 & 51182 & 190 & 0.22 \\
\multicolumn{5}{l}{\textbf{Laboratory Abnormality}}\\
High white blood count & 79413 & 47311 & 11209 & 14.11 \\
Low white blood count & 83847 & 49932 & 8673 & 10.34 \\
High absolute neutrophil count & 80692 & 47861 & 9873 & 12.24 \\
Low absolute neutrophil count & 86130 & 50754 & 6581 & 7.64 \\
High bands & 82411 & 48715 & 11817 & 14.34 \\
High lymphocyte & 85942 & 50567 & 4052 & 4.71 \\
Low lymphocyte & 81840 & 48854 & 12300 & 15.03 \\
High hemoglobin & 84190 & 49804 & 3845 & 4.57 \\
Low hemoglobin & 76558 & 46493 & 17200 & 22.47 \\
High mean corpuscular volume & 84106 & 50126 & 6249 & 7.43 \\
Low mean corpuscular volume & 84325 & 49823 & 4297 & 5.10 \\
High reticulocyte count & 86490 & 50878 & 3671 & 4.24 \\
Low reticulocyte count & 86829 & 50884 & 2783 & 3.21 \\
High platelet & 82802 & 49347 & 10999 & 13.28 \\
Low platelet & 80032 & 48114 & 11532 & 14.41 \\
High immature platelet fraction & 85633 & 50531 & 5172 & 6.04 \\
Low immature platelet fraction & 86820 & 50895 & 2372 & 2.73 \\
High mean platelet volume & 85932 & 50628 & 4504 & 5.24 \\
Low mean platelet volume & 83497 & 49309 & 7682 & 9.20 \\
High fibrinogen & 86881 & 50852 & 2074 & 2.39 \\
Low fibrinogen & 84828 & 49769 & 1568 & 1.85 \\
High partial thromboplastin time & 84523 & 49896 & 3330 & 3.94 \\
High international normalized ratio & 81124 & 48207 & 6544 & 8.07 \\
High sodium & 80268 & 47701 & 9034 & 11.25 \\
Low sodium & 84152 & 49895 & 7330 & 8.71 \\
High potassium & 83238 & 49323 & 7071 & 8.49 \\
Low potassium & 79405 & 47367 & 13023 & 16.40 \\
High glucose & 76531 & 45675 & 10759 & 14.06 \\
Low glucose & 85587 & 50554 & 3735 & 4.36 \\
High creatinine & 84100 & 49855 & 4814 & 5.72 \\
High urea & 86106 & 50794 & 3135 & 3.64 \\
Low albumin & 84353 & 49870 & 10279 & 12.19 \\
High alanine transaminase & 83472 & 49667 & 8080 & 9.68 \\
High aspartate aminotransferase & 84283 & 49930 & 6830 & 8.10 \\
High lactate dehydrogenase & 86601 & 50809 & 2330 & 2.69 \\
High bilirubin & 83640 & 49416 & 4785 & 5.72 \\
High cholesterol & 87300 & 51131 & 537 & 0.62 \\
High triglyceride & 86804 & 50825 & 2349 & 2.71 \\
High ferritin & 86385 & 50637 & 3089 & 3.58 \\
High creatinine kinase & 87037 & 50919 & 675 & 0.78 \\
High C reactive protein & 81869 & 48546 & 10987 & 13.42 \\
High erythrocyte sedimentation rate & 86470 & 50702 & 2748 & 3.18 \\
Low PaO2 & 86622 & 50790 & 2492 & 2.88 \\
Low SpO2 & 60215 & 36121 & 20632 & 34.26 \\
\multicolumn{5}{l}{\textbf{Medications}}\\
Any antibacterial & 55021 & 35011 & 17411 & 31.64 \\
Any antifungal & 86729 & 51130 & 2142 & 2.47 \\
Any chemotherapy & 83359 & 50997 & 3092 & 3.71 \\
Any antiepileptics & 78305 & 48262 & 6578 & 8.40 \\
Any glucocorticoid & 67347 & 40508 & 14145 & 21.00 \\
Dexamethasone & 70377 & 41571 & 9815 & 13.95 \\
Any opioid & 59884 & 34993 & 16019 & 26.75 \\
Morphine & 67581 & 39484 & 12160 & 17.99 \\
Fentanyl & 69306 & 40392 & 12485 & 18.01 \\
Any inotrope & 83893 & 49299 & 1861 & 2.22 \\
\multicolumn{5}{l}{\textbf{Clinical Outcomes}}\\
Long length of stay ($\geq$ 7 days) & 87565 & 51242 & 19422 & 22.18 \\
Readmission within 30 days & 86395 & 50586 & 14807 & 17.14 \\
Mortality & 87447 & 51182 & 514 & 0.59 \\
\multicolumn{5}{l}{\textbf{MIMIC}}\\
\multicolumn{5}{l}{\textbf{Laboratory Abnormality}}\\
High hemoglobin & 58403 & 43968 & 88 & 0.15 \\
Low hemoglobin & 35499 & 28992 & 15397 & 43.37 \\
High platelet & 56924 & 43179 & 3169 & 5.57 \\
Low platelet & 52218 & 39892 & 8889 & 17.02 \\
High sodium & 57390 & 43241 & 3900 & 6.80 \\
Low sodium & 54759 & 41526 & 8457 & 15.44 \\
High potassium & 55915 & 42348 & 6168 & 11.03 \\
Low potassium & 56570 & 42678 & 8118 & 14.35 \\
High glucose & 39084 & 30204 & 14428 & 36.92 \\
Low glucose & 57967 & 43715 & 3732 & 6.44 \\
\multicolumn{5}{l}{\textbf{Clinical Outcomes}}\\
Long length of stay ($\geq$ 7 days) & 58513 & 44055 & 17218 & 29.43 \\
Readmission within 30 days & 58512 & 44055 & 3143 & 5.37 \\
Mortality & 58513 & 44055 & 1741 & 2.98 \\
\bottomrule
\end{longtable}
\par\vspace{0.4em}
{\footnotesize\noindent  Abbreviations: SickKids -- The Hospital for Sick Children; MIMIC -- Medical Information Mart for Intensive Care.}

\begin{table}[H]
\centering
\small
\caption*{Supplementary Table S6. Tokenization and Vocabulary Statistics}
\begin{tabular}{lllrrr}
\toprule
Event & Time & Workflow & Vocabulary Size & Total Tokens (Pretraining)* & Mean Tokens per Patient \\
\midrule
Joint & Positions & No & 25,263 & 169,323,884 & 83.51 \\
Joint & Positions & Yes & 35,373 & 225,713,027 & 111.32 \\
Joint & Tokens & No & 25,322 & 197,021,149 & 214.14 \\
Joint & Tokens & Yes & 35,432 & 256,387,844 & 278.60 \\
Factorized & Positions & No & 10,889 & 331,800,377 & 163.64 \\
Factorized & Positions & Yes & 11,758 & 448,086,342 & 221.00 \\
Factorized & Tokens & No & 10,948 & 359,497,647 & 390.73 \\
Factorized & Tokens & Yes & 11,817 & 478,761,166 & 520.24 \\
\bottomrule
\end{tabular}
\par\vspace{0.4em}
{\footnotesize\noindent  *Total tokens used for pretraining include training and validation sets with temporal cutoffs.}
\end{table}

\begin{table}[H]
\centering
\small
\caption*{Supplementary Table S7. Pretraining Compute and Tokens Seen}
\begin{tabular}{lllrrr}
\toprule
Event & Time & Workflow & Total FLOPs (Pretraining) & Tokens Seen (Pretraining) & Steps for 5 Epochs \\
\midrule
Joint & Positions & No & $5.51 \times 10^{17}$ & 680,615,416 & 11,965 \\
Joint & Positions & Yes & $7.70 \times 10^{17}$ & 899,291,143 & 16,200 \\
Joint & Tokens & No & $6.37 \times 10^{17}$ & 785,851,344 & 14,055 \\
Joint & Tokens & Yes & $8.71 \times 10^{17}$ & 1,015,924,901 & 18,555 \\
Factorized & Positions & No & $9.61 \times 10^{17}$ & 1,291,693,160 & 24,300 \\
Factorized & Positions & Yes & $1.29 \times 10^{18}$ & 1,719,058,463 & 33,095 \\
Factorized & Tokens & No & $1.04 \times 10^{18}$ & 1,390,730,204 & 26,360 \\
Factorized & Tokens & Yes & $1.37 \times 10^{18}$ & 1,831,385,376 & 35,495 \\
\bottomrule
\end{tabular}
\par\vspace{0.4em}
{\footnotesize\noindent  Differences in total FLOPs, number of tokens seen, and training steps arise from differences in effective sequence length across tokenization strategies. Abbreviation: FLOPs -- floating-point operations.}
\end{table}

\begin{table}[H]
\centering
\small
\caption*{Supplementary Table S8. Main Experiment Full-Shot Performance}
\begin{tabular}{llll}
\toprule
Event Encoding & Time Encoding & Workflow & Mean AUROC* \\
\midrule
Factorized & Time Positions & No & 0.872 \\
Factorized & Time Positions & Yes & 0.881 \\
Joint & Time Positions & No & 0.882 \\
Joint & Time Positions & Yes & 0.886 \\
Factorized & Time Tokens & No & 0.865 \\
Factorized & Time Tokens & Yes & 0.874 \\
Joint & Time Tokens & No & 0.876 \\
Joint & Time Tokens & Yes & 0.880 \\
\bottomrule
\end{tabular}
\par\vspace{0.4em}
{\footnotesize\noindent  * Mean AUROC across 74 clinical prediction tasks. \\ Abbreviation: AUROC -- area under the receiver operating characteristics curve.}
\end{table}

\scriptsize
\begin{longtable}{lcccccccc}
\caption*{Supplementary Table S9. Main Experiment Full-Shot AUROC by Task and Tokenization Condition}\\
\toprule
Task Name & Pos/F/No & Pos/F/Yes & Pos/J/No & Pos/J/Yes & Tok/F/No & Tok/F/Yes & Tok/J/No & Tok/J/Yes \\
\midrule
\endfirsthead
\toprule
Task Name & Pos/F/No & Pos/F/Yes & Pos/J/No & Pos/J/Yes & Tok/F/No & Tok/F/Yes & Tok/J/No & Tok/J/Yes \\
\midrule
\endhead
Platelet transfusion & 0.961 & 0.962 & 0.961 & 0.961 & 0.952 & 0.953 & 0.958 & 0.960 \\
Red cell transfusion & 0.913 & 0.912 & 0.922 & 0.922 & 0.907 & 0.912 & 0.917 & 0.915 \\
Invasive intubation & 0.928 & 0.933 & 0.936 & 0.931 & 0.920 & 0.931 & 0.936 & 0.934 \\
Gastrostomy tube & 0.914 & 0.902 & 0.925 & 0.936 & 0.921 & 0.905 & 0.920 & 0.910 \\
Echocardiogram & 0.869 & 0.892 & 0.874 & 0.897 & 0.866 & 0.883 & 0.871 & 0.882 \\
Pulmonary function test & 0.957 & 0.955 & 0.965 & 0.970 & 0.960 & 0.948 & 0.966 & 0.953 \\
Lumbar puncture & 0.936 & 0.952 & 0.948 & 0.949 & 0.940 & 0.949 & 0.948 & 0.946 \\
Surgery & 0.889 & 0.890 & 0.886 & 0.891 & 0.880 & 0.887 & 0.885 & 0.891 \\
Interventional radiology & 0.855 & 0.869 & 0.859 & 0.872 & 0.848 & 0.866 & 0.850 & 0.867 \\
Plain radiography chest & 0.808 & 0.832 & 0.821 & 0.822 & 0.803 & 0.816 & 0.812 & 0.823 \\
Ultrasound abdomen & 0.829 & 0.859 & 0.843 & 0.857 & 0.824 & 0.843 & 0.831 & 0.850 \\
Computerized tomography chest & 0.880 & 0.894 & 0.877 & 0.889 & 0.872 & 0.882 & 0.869 & 0.886 \\
Computerized tomography abdomen & 0.883 & 0.891 & 0.891 & 0.894 & 0.878 & 0.887 & 0.880 & 0.882 \\
Computerized tomography head & 0.908 & 0.919 & 0.917 & 0.922 & 0.903 & 0.909 & 0.912 & 0.917 \\
MRI head & 0.911 & 0.915 & 0.915 & 0.913 & 0.901 & 0.914 & 0.902 & 0.914 \\
MRI whole body & 0.934 & 0.921 & 0.934 & 0.917 & 0.901 & 0.874 & 0.900 & 0.914 \\
PET & 0.911 & 0.900 & 0.909 & 0.914 & 0.851 & 0.874 & 0.894 & 0.910 \\
High white blood count & 0.800 & 0.822 & 0.829 & 0.835 & 0.792 & 0.811 & 0.818 & 0.821 \\
Low white blood count & 0.896 & 0.899 & 0.921 & 0.912 & 0.896 & 0.895 & 0.911 & 0.907 \\
High absolute neutrophil count & 0.819 & 0.828 & 0.839 & 0.836 & 0.807 & 0.819 & 0.827 & 0.824 \\
Low absolute neutrophil count & 0.900 & 0.905 & 0.919 & 0.914 & 0.898 & 0.901 & 0.912 & 0.910 \\
High bands & 0.857 & 0.871 & 0.868 & 0.873 & 0.851 & 0.864 & 0.861 & 0.868 \\
High lymphocyte & 0.832 & 0.841 & 0.850 & 0.851 & 0.825 & 0.831 & 0.844 & 0.841 \\
Low lymphocyte & 0.876 & 0.882 & 0.889 & 0.888 & 0.872 & 0.879 & 0.880 & 0.885 \\
High hemoglobin & 0.825 & 0.840 & 0.842 & 0.847 & 0.819 & 0.830 & 0.829 & 0.843 \\
Low hemoglobin & 0.868 & 0.881 & 0.882 & 0.889 & 0.863 & 0.877 & 0.877 & 0.885 \\
High mean corpuscular volume & 0.865 & 0.868 & 0.886 & 0.878 & 0.853 & 0.862 & 0.875 & 0.869 \\
Low mean corpuscular volume & 0.812 & 0.826 & 0.847 & 0.848 & 0.791 & 0.814 & 0.822 & 0.824 \\
High reticulocyte count & 0.891 & 0.902 & 0.905 & 0.907 & 0.885 & 0.899 & 0.899 & 0.903 \\
Low reticulocyte count & 0.867 & 0.882 & 0.876 & 0.889 & 0.855 & 0.874 & 0.877 & 0.875 \\
High platelet & 0.830 & 0.845 & 0.847 & 0.848 & 0.824 & 0.838 & 0.842 & 0.843 \\
Low platelet & 0.872 & 0.876 & 0.884 & 0.884 & 0.867 & 0.871 & 0.880 & 0.882 \\
High immature platelet fraction & 0.872 & 0.887 & 0.882 & 0.887 & 0.865 & 0.877 & 0.884 & 0.884 \\
Low immature platelet fraction & 0.859 & 0.875 & 0.871 & 0.882 & 0.850 & 0.853 & 0.870 & 0.877 \\
High mean platelet volume & 0.894 & 0.900 & 0.906 & 0.900 & 0.882 & 0.889 & 0.901 & 0.900 \\
Low mean platelet volume & 0.786 & 0.795 & 0.796 & 0.806 & 0.774 & 0.785 & 0.793 & 0.792 \\
High fibrinogen & 0.894 & 0.887 & 0.892 & 0.893 & 0.886 & 0.887 & 0.890 & 0.886 \\
Low fibrinogen & 0.893 & 0.896 & 0.908 & 0.902 & 0.884 & 0.890 & 0.896 & 0.896 \\
High partial thromboplastin time & 0.890 & 0.897 & 0.898 & 0.897 & 0.886 & 0.888 & 0.895 & 0.897 \\
High international normalized ratio & 0.866 & 0.873 & 0.886 & 0.890 & 0.863 & 0.879 & 0.880 & 0.881 \\
High sodium & 0.828 & 0.837 & 0.852 & 0.848 & 0.828 & 0.842 & 0.842 & 0.843 \\
Low sodium & 0.831 & 0.841 & 0.850 & 0.850 & 0.828 & 0.834 & 0.839 & 0.842 \\
High potassium & 0.826 & 0.834 & 0.835 & 0.834 & 0.818 & 0.832 & 0.827 & 0.836 \\
Low potassium & 0.855 & 0.863 & 0.870 & 0.867 & 0.851 & 0.861 & 0.859 & 0.865 \\
High glucose & 0.829 & 0.840 & 0.839 & 0.846 & 0.822 & 0.835 & 0.832 & 0.843 \\
Low glucose & 0.863 & 0.869 & 0.881 & 0.879 & 0.861 & 0.869 & 0.872 & 0.873 \\
High creatinine & 0.886 & 0.893 & 0.905 & 0.899 & 0.874 & 0.885 & 0.888 & 0.893 \\
High urea & 0.916 & 0.924 & 0.933 & 0.928 & 0.912 & 0.919 & 0.926 & 0.924 \\
Low albumin & 0.879 & 0.888 & 0.885 & 0.888 & 0.874 & 0.884 & 0.880 & 0.888 \\
High alanine transaminase & 0.859 & 0.870 & 0.879 & 0.882 & 0.858 & 0.866 & 0.874 & 0.868 \\
High aspartate aminotransferase & 0.859 & 0.871 & 0.880 & 0.883 & 0.852 & 0.863 & 0.875 & 0.872 \\
High lactate dehydrogenase & 0.890 & 0.918 & 0.898 & 0.912 & 0.893 & 0.899 & 0.894 & 0.900 \\
High bilirubin & 0.884 & 0.894 & 0.896 & 0.899 & 0.878 & 0.891 & 0.890 & 0.896 \\
High cholesterol & 0.900 & 0.904 & 0.887 & 0.905 & 0.858 & 0.871 & 0.876 & 0.886 \\
High triglyceride & 0.830 & 0.847 & 0.850 & 0.863 & 0.840 & 0.854 & 0.843 & 0.853 \\
High ferritin & 0.865 & 0.888 & 0.875 & 0.889 & 0.859 & 0.874 & 0.876 & 0.883 \\
High creatinine kinase & 0.901 & 0.904 & 0.909 & 0.923 & 0.872 & 0.895 & 0.897 & 0.910 \\
High C-reactive protein & 0.844 & 0.850 & 0.851 & 0.859 & 0.835 & 0.854 & 0.842 & 0.851 \\
High erythrocyte sedimentation rate & 0.910 & 0.913 & 0.922 & 0.915 & 0.907 & 0.910 & 0.917 & 0.903 \\
Low PaO2 & 0.954 & 0.951 & 0.956 & 0.952 & 0.947 & 0.953 & 0.950 & 0.950 \\
Low SpO2 & 0.808 & 0.814 & 0.814 & 0.824 & 0.802 & 0.811 & 0.808 & 0.823 \\
Any antibacterial & 0.858 & 0.874 & 0.866 & 0.879 & 0.849 & 0.878 & 0.858 & 0.875 \\
Any antifungal & 0.936 & 0.943 & 0.939 & 0.946 & 0.928 & 0.948 & 0.937 & 0.944 \\
Any chemotherapy & 0.967 & 0.972 & 0.973 & 0.970 & 0.962 & 0.967 & 0.961 & 0.966 \\
Any antiepileptics & 0.850 & 0.855 & 0.852 & 0.862 & 0.848 & 0.860 & 0.845 & 0.860 \\
Any glucocorticoid & 0.820 & 0.834 & 0.824 & 0.834 & 0.818 & 0.832 & 0.823 & 0.837 \\
Dexamethasone & 0.823 & 0.830 & 0.823 & 0.833 & 0.820 & 0.832 & 0.825 & 0.839 \\
Any opioid & 0.856 & 0.870 & 0.858 & 0.871 & 0.853 & 0.868 & 0.859 & 0.872 \\
Morphine & 0.839 & 0.867 & 0.837 & 0.863 & 0.834 & 0.862 & 0.841 & 0.863 \\
Fentanyl & 0.853 & 0.856 & 0.852 & 0.859 & 0.846 & 0.853 & 0.852 & 0.855 \\
Any inotrope & 0.915 & 0.916 & 0.932 & 0.919 & 0.903 & 0.912 & 0.913 & 0.915 \\
Long length of stay (>= 7 days) & 0.812 & 0.825 & 0.822 & 0.828 & 0.817 & 0.823 & 0.822 & 0.826 \\
Readmission within 30 days & 0.794 & 0.798 & 0.796 & 0.792 & 0.794 & 0.795 & 0.795 & 0.802 \\
Mortality & 0.940 & 0.941 & 0.946 & 0.939 & 0.925 & 0.929 & 0.943 & 0.940 \\
\bottomrule
\end{longtable}
\par\vspace{0.4em}
{\footnotesize\noindent  Abbreviation: AUROC -- area under the receiver operating characteristics curve; Pos -- positions; Tok -- tokens; F -- factorized; J -- joint.}

\begin{table}[H]
\centering
\small
\caption*{Supplementary Table S10. Main Experiment Fixed-Effect Estimates}
\begin{tabular}{llll}
\toprule
Effect & $\beta$ (AUROC) & 95\% CI & p value \\
\midrule
Intercept* & 0.870 & [0.860, 0.880] & $<0.001$ \\
Time Encoding: Time-Positions vs. Time-Tokens & 0.007 & [0.005, 0.008] & $<0.001$ \\
Event Encoding: Joint vs Factorized & 0.008 & [0.007, 0.009] & $<0.001$ \\
Workflow: Yes vs. No & 0.007 & [0.006, 0.008] & $<0.001$ \\
\bottomrule
\end{tabular}
\par\vspace{0.4em}
{\footnotesize\noindent  Fixed-effect estimates from a linear mixed-effects model where task-level AUROC is the outcome across 74 clinical prediction tasks. The model included time encoding, event encoding, and workflow as fixed effects, with task included as a random intercept. p-values were computed using a Wald t-distribution approximation.\\ **Intercept reflects Time-Positions, Factorized and No Workflow. \\ Abbreviations: AUROC -- area under the receiver operating characteristics curve; CI -- confidence interval.}
\end{table}

\subsection*{Supplementary Figure S2. Effect of Tokenization Design Choices on Sample Efficiency}
\begin{figure}[H]
\centering
\IfFileExists{S_few_shot_performance.pdf}{%
  \includegraphics[width=\textwidth]{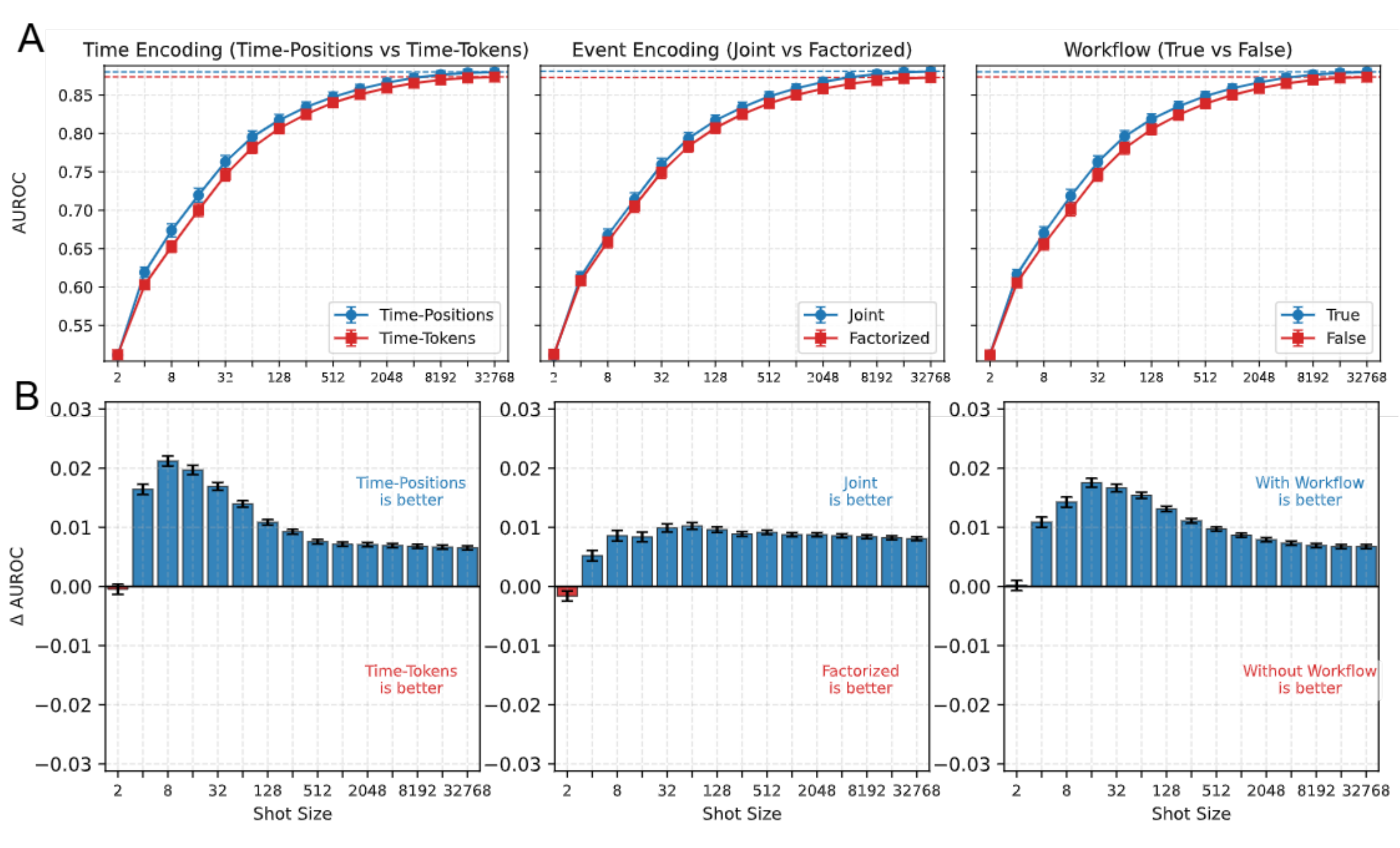}
}{%
  \fbox{\parbox{0.95\textwidth}{\centering Placeholder for Supplementary Figure S2\\
  Expected file: \texttt{SF2\_sample\_efficiency.pdf} (not currently present in workspace).}}
}
\caption{Effect of Tokenization Design Choices on Sample Efficiency. (A) Mean AUROC as a function of the number of labeled training examples (shots) for time encoding (Time-Positions vs Time-Tokens), event encoding (Joint vs Factorized), and workflow (With vs Without). Error bars indicate $\pm$1 standard error. Horizontal dashed lines denote average full-shot AUROC. (B) Mean AUROC differences between paired tokenization strategies at each shot size. Error bars denote $\pm$1 standard error. Abbreviation: AUROC -- area under the receiver operating characteristics curve.}
\label{fig:supp_s2}
\end{figure}

\begin{table}[H]
\centering
\small
\caption*{Supplementary Table S11. Sample Efficiency Evaluation Fixed-Effect and Interaction Estimates}
\begin{tabular}{llll}
\toprule
Effect & $\beta$ (AUROC) & 95\% CI & p value \\
\midrule
\multicolumn{4}{l}{\textbf{Main effects (at 32 shot)}} \\
Time Encoding: Time Positions vs Time Tokens & 0.006 & [0.005, 0.007] & $<0.001$ \\
Event Encoding: Joint vs. Factorized & 0.008 & [0.007, 0.009] & $<0.001$ \\
Workflow: Yes vs No & 0.007 & [0.006, 0.008] & $<0.001$ \\
\multicolumn{4}{l}{\textbf{Interactions with $\log_2$(shot size)}} \\
Time encoding $\times$ $\log_2$(shot size) & 0.001 & [0.000, 0.001] & 0.012 \\
Event encoding $\times$ $\log_2$(shot size) & 0.000 & [-0.000, 0.001] & 0.266 \\
Workflow $\times$ $\log_2$(shot size) & -0.000 & [-0.001, 0.000] & 0.161 \\
\bottomrule
\end{tabular}
\par\vspace{0.4em}
{\footnotesize\noindent  Fixed-effect estimates from a linear mixed-effects model where task-level AUROC is the outcome across 74 clinical prediction tasks. The model included time encoding, event encoding, workflow, log-transformed shot size (centered at 32 examples), and their interactions as fixed effects, with task included as a random intercept. Reference levels are Time-Positions, Factorized, and No Workflow. p-values were computed using a Wald t-distribution approximation. Abbreviations: AUROC -- area under the receiver operating characteristics curve; CI -- confidence interval.}
\end{table}

\begin{table}[H]
\centering
\small
\caption*{Supplementary Table S12. Event Encoding Ablation Fixed-Effect and Interaction Estimates}
\begin{tabular}{llll}
\toprule
Effect & $\beta$ (AUROC) & 95\% CI & p value \\
\midrule
\multicolumn{4}{l}{\textbf{Main effects}} \\
Factorized (vs Joint) & -0.001 & [-0.003, 0.001] & 0.230 \\
Attributes (vs Code only) & 0.009 & [0.007, 0.010] & $<0.001$ \\
Workflow (vs Code only) & 0.007 & [0.005, 0.009] & $<0.001$ \\
Full (vs Code only) & 0.013 & [0.012, 0.015] & $<0.001$ \\
Fixed event (vs Fixed length) & 0.000 & [-0.002, 0.002] & 0.935 \\
\multicolumn{4}{l}{\textbf{Key interactions}} \\
Factorized $\times$ Attributes & -0.009 & [-0.011, -0.006] & $<0.001$ \\
Factorized $\times$ Workflow & 0.002 & [-0.001, 0.005] & 0.188 \\
Factorized $\times$ Full & -0.005 & [-0.008, -0.002] & $<0.001$ \\
\bottomrule
\end{tabular}
\par\vspace{0.4em}
{\footnotesize\noindent  Fixed-effect estimates from a linear mixed-effects model where task-level AUROC is the outcome across 74 clinical prediction tasks. The model included event encoding (Joint vs Factorized), information content (Code only, +Attributes, +Workflow, Full), sequence length regime (Fixed-length vs Fixed-event), and corresponding interactions as fixed effects, with task included as a random intercept. Reference levels are Joint encoding, Code-only information content, and Fixed-length regime. p-values were computed using a Wald t-distribution approximation. All interaction terms not shown in the table were non-significant (p > 0.958). Abbreviations: AUROC -- area under the receiver operating characteristics curve; CI -- confidence interval.}
\end{table}

\begin{table}[H]
\centering
\small
\caption*{Supplementary Table S13. Time Encoding Ablation Fixed-Effect Estimates}
\begin{tabular}{llll}
\toprule
Time encoding (vs Order only) & $\beta$ (AUROC) & 95\% CI & p value \\
\midrule
Time Positions (RoPE) & 0.003 & [0.001, 0.004] & $<0.001$ \\
Positions + Scalar & 0.003 & [0.001, 0.004] & $<0.001$ \\
Time Tokens & -0.003 & [-0.005, -0.002] & $<0.001$ \\
\bottomrule
\end{tabular}
\par\vspace{0.4em}
{\footnotesize\noindent  Fixed-effect estimates from a linear mixed-effects model where task-level AUROC is the outcome across 74 clinical prediction tasks. The model included time encoding strategy (Order-only, Time-Positions, Time-Tokens, Positions + Scalar) as a fixed effect, with task included as a random intercept. p-values were computed using a Wald t-distribution approximation. Abbreviations: AUROC -- area under the receiver operating characteristics curve; CI -- confidence interval.}
\end{table}

\begin{table}[H]
\centering
\small
\caption*{Supplementary Table S14. Transfer Performance on MIMIC by Tokenization Condition}
\begin{tabular}{llll}
\toprule
Event Encoding & Time Encoding & Workflow & Mean AUROC* \\
\midrule
Factorized & Time-Positions & No & 0.808 \\
Factorized & Time Positions & Yes & 0.806 \\
Joint & Time Positions & No & 0.815 \\
Joint & Time Positions & Yes & 0.813 \\
Factorized & Time Tokens & No & 0.804 \\
Factorized & Time Tokens & Yes & 0.805 \\
Joint & Time Tokens & No & 0.811 \\
Joint & Time Tokens & Yes & 0.815 \\
\bottomrule
\end{tabular}
\par\vspace{0.4em}
{\footnotesize\noindent  * Mean AUROC for 13 MIMIC clinical prediction tasks using linear probes trained on frozen foundation models pretrained on SickKids. Abbreviation: AUROC -- area under the receiver operating characteristics curve.}
\end{table}

\small
\begin{longtable}{lcccccccc}
\caption*{Supplementary Table S15. Transfer AUROC on MIMIC by Task and Tokenization Condition}\\
\toprule
Task Name & Pos/F/No & Pos/F/Yes & Pos/J/No & Pos/J/Yes & Toks/F/No & Toks/F/Yes & Toks/J/No & Toks/J/Yes \\
\midrule
\endfirsthead
\toprule
Task Name & Pos/F/No & Pos/F/Yes & Pos/J/No & Pos/J/Yes & Toks/F/No & Toks/F/Yes & Toks/J/No & Toks/J/Yes \\
\midrule
\endhead
High hemoglobin & 0.677 & 0.721 & 0.723 & 0.692 & 0.704 & 0.696 & 0.670 & 0.720 \\
Low hemoglobin & 0.853 & 0.851 & 0.849 & 0.851 & 0.856 & 0.857 & 0.856 & 0.855 \\
High platelet & 0.765 & 0.754 & 0.775 & 0.764 & 0.757 & 0.761 & 0.777 & 0.776 \\
Low platelet & 0.806 & 0.796 & 0.814 & 0.814 & 0.804 & 0.797 & 0.816 & 0.811 \\
High sodium & 0.795 & 0.788 & 0.819 & 0.816 & 0.789 & 0.795 & 0.809 & 0.810 \\
Low sodium & 0.804 & 0.799 & 0.816 & 0.815 & 0.792 & 0.796 & 0.815 & 0.812 \\
High potassium & 0.814 & 0.808 & 0.817 & 0.821 & 0.802 & 0.802 & 0.819 & 0.817 \\
Low potassium & 0.809 & 0.800 & 0.819 & 0.816 & 0.806 & 0.808 & 0.820 & 0.819 \\
High glucose & 0.921 & 0.921 & 0.920 & 0.920 & 0.925 & 0.924 & 0.922 & 0.920 \\
Low glucose & 0.766 & 0.763 & 0.764 & 0.770 & 0.758 & 0.768 & 0.763 & 0.773 \\
Long length of stay (>= 7 days) & 0.794 & 0.791 & 0.785 & 0.789 & 0.789 & 0.788 & 0.782 & 0.789 \\
Readmission within 30 days & 0.815 & 0.819 & 0.816 & 0.814 & 0.809 & 0.813 & 0.820 & 0.818 \\
Mortality & 0.887 & 0.871 & 0.883 & 0.889 & 0.859 & 0.861 & 0.880 & 0.881 \\
\bottomrule
\end{longtable}
\par\vspace{0.4em}
{\footnotesize\noindent  Abbreviation: AUROC -- area under the receiver operating characteristics curve; Pos -- positions; Toks -- tokens; Factor -- factorized.}

\begin{table}[H]
\centering
\small
\caption*{Supplementary Table S16. External Evaluation (MIMIC) Fixed-Effect Estimates}
\begin{tabular}{llll}
\toprule
Effect & $\beta$ (AUROC) & 95\% CI & p value \\
\midrule
Intercept* & 0.806 & [0.777, 0.835] & $<0.001$ \\
Time Encoding: Time Positions vs Time Tokens & 0.002 & [-0.001, 0.005] & 0.271 \\
Event Encoding: Joint vs Factorized & 0.008 & [0.005, 0.011] & $<0.001$ \\
Workflow: Yes vs No & 0.000 & [-0.003, 0.004] & 0.803 \\
\bottomrule
\end{tabular}
\par\vspace{0.4em}
{\footnotesize\noindent  Fixed-effect estimates from a linear mixed-effects model where task-level AUROC is the outcome across 13 MIMIC clinical prediction tasks in the external evaluation of the SickKids foundation model. The model included time encoding, event encoding, and workflow as fixed effects, with task included as a random intercept. p-values were computed using a Wald t-distribution approximation.}
\par\vspace{0.4em}
{\footnotesize\noindent  *Intercept reflects Time-Positions, Factorized and No Workflow. Abbreviations: AUROC -- area under the receiver operating characteristics curve; CI -- confidence interval.}
\end{table}

\begin{table}[H]
\centering
\small
\caption*{Supplementary Table S17. MIMIC Reference Performance by Tokenization Condition}
\begin{tabular}{lll}
\toprule
Time Encoding & Event Encoding & Mean AUROC* \\
\midrule
Time Positions & Factorized & 0.839 \\
Time Positions & Joint & 0.842 \\
Time Tokens & Factorized & 0.836 \\
Time Tokens & Joint & 0.836 \\
\bottomrule
\end{tabular}
\par\vspace{0.4em}
{\footnotesize\noindent  *Mean AUROC across 13 MIMIC clinical prediction tasks for models pretrained directly on MIMIC using a MIMIC-derived vocabulary. These results represent an upper-bound reference for transfer experiments, where models pretrained at SickKids are evaluated on MIMIC using a fixed source vocabulary. Abbreviation: AUROC -- area under the receiver operating characteristics curve.}
\end{table}

\small
\begin{longtable}{lcccc}
\caption*{Supplementary Table S18. MIMIC Reference AUROC by Task and Tokenization Condition}\\
\toprule
Task Name & Pos/Factor & Pos/Joint & Toks/Factor & Toks/Joint \\
\midrule
\endfirsthead
\toprule
Task Name & Pos/Factor & Pos/Joint & Toks/Factor & Toks/Joint \\
\midrule
\endhead
High hemoglobin & 0.649 & 0.659 & 0.688 & 0.629 \\
Low hemoglobin & 0.879 & 0.876 & 0.869 & 0.874 \\
High platelet & 0.820 & 0.834 & 0.811 & 0.824 \\
Low platelet & 0.841 & 0.850 & 0.830 & 0.840 \\
High sodium & 0.833 & 0.844 & 0.821 & 0.833 \\
Low sodium & 0.835 & 0.837 & 0.832 & 0.838 \\
High potassium & 0.850 & 0.848 & 0.845 & 0.847 \\
Low potassium & 0.832 & 0.835 & 0.836 & 0.837 \\
High glucose & 0.938 & 0.935 & 0.935 & 0.938 \\
Low glucose & 0.816 & 0.815 & 0.802 & 0.805 \\
Long length of stay (>= 7 days) & 0.840 & 0.833 & 0.833 & 0.828 \\
Readmission within 30 days & 0.850 & 0.852 & 0.842 & 0.843 \\
Mortality & 0.926 & 0.932 & 0.928 & 0.930 \\
\bottomrule
\end{longtable}
{\footnotesize\noindent  Abbreviation: AUROC -- area under the receiver operating characteristics curve; Pos -- positions; Toks -- tokens; Factor -- factorized.}

\end{document}